\documentclass{article}

\PassOptionsToPackage{numbers}{natbib} 

\usepackage[main, final]{neurips_2026}

\usepackage[utf8]{inputenc} 
\usepackage[T1]{fontenc}    
\usepackage{hyperref}       
\usepackage{url}            
\usepackage{booktabs}       
\usepackage{amsfonts}       
\usepackage{nicefrac}       
\usepackage{microtype}      
\usepackage{xcolor}      
\usepackage{multirow}
\usepackage{colortbl}
\usepackage{algorithm}
\usepackage{algpseudocode}
\usepackage{graphicx}
\usepackage{enumitem}
\usepackage{amsmath}
\usepackage{svg}
\usepackage{subcaption}
\usepackage{wrapfig}
\usepackage{fontawesome5}
\definecolor{githublink}{HTML}{6765A8}

\definecolor{bestcolor}{RGB}{246,221,221}      
\definecolor{secondcolor}{RGB}{247,239,205}    
\definecolor{groupcolorA}{RGB}{230,230,230}
\definecolor{groupcolorB}{RGB}{228,208,176}
\definecolor{improvecolor}{RGB}{0,106,78}       
\definecolor{degradecolor}{RGB}{139,35,35}       

\title{HALMES: Knowing When to Intervene in LVLM Hallucination Mitigation}

\author{%
  Ruining Hu$^{1}$\thanks{Equal contribution.} \quad
  Jiaqi Lu$^{1}$\footnotemark[1] \quad
  Xiao Liu$^{1}$ \quad
  Ying Shen$^{1}$\thanks{Corresponding author.} \quad
  Yu Wang$^{1}$ \quad
  Lin Zhang$^{1,2}$ \\
  $^{1}$School of Computer Science and Technology, Tongji University \\
  $^{2}$IEIT SYSTEMS (Beijing) Co., Ltd \\
  \texttt{\{2534075,2512091,2534063,yingshen,cslinzhang\}@tongji.edu.cn} \\
  \texttt{yuwangtj@yeah.net}
  \\[4pt]
  {%
    \hypersetup{pdfborder={0 0 0}}%
    \href{https://github.com/slp-tongji/Halmes}{%
        \textcolor{black}{\faGithub}\ %
        \textcolor{githublink}{\texttt{https://github.com/slp-tongji/Halmes}}%
    }%
  }
}

\begin{document}

\maketitle

\begin{figure*}[htbp]
    \centering
    \includegraphics[width=0.8\textwidth]{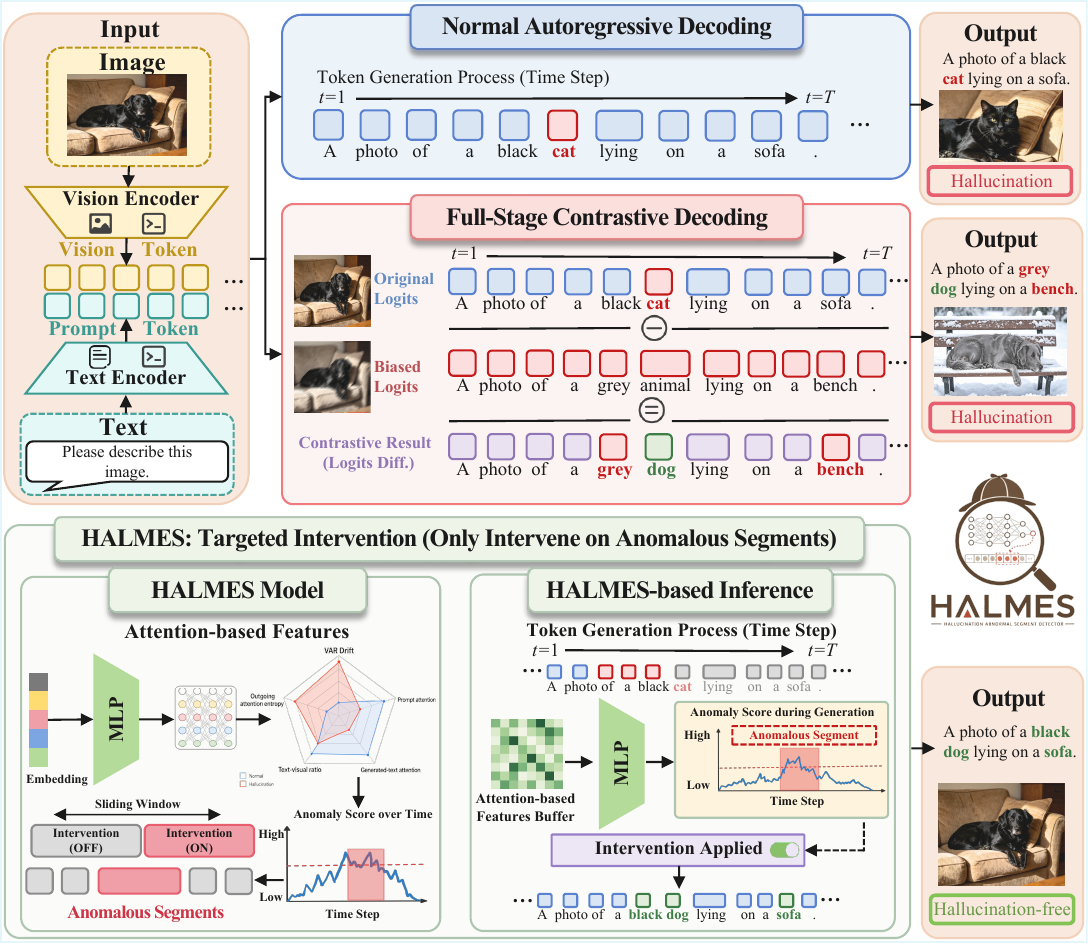}
    \caption{Overview of HALMES. By detecting anomalous token segments from internal attention features, HALMES enables early hallucination warning and selectively triggers contrastive decoding, transforming full-stage indiscriminate intervention into on-demand targeted correction.}
    \label{fig:framework}
\end{figure*}

\begin{abstract}
Large Vision-Language Models (LVLMs) have demonstrated strong capabilities in multimodal tasks. Nevertheless, their outputs remain prone to object hallucinations that are inconsistent with visual content. Contrastive decoding, as a training-free inference-time mitigation approach, typically intervenes in the output distribution at every decoding step. However, such indiscriminate full-stage intervention may disrupt originally correct generations and even induce new hallucinations. To address this issue, we first analyze the relationship between internal attention features and hallucinated token segments in LVLMs, revealing that hallucination-related abnormalities progressively emerge in preceding token segments. Based on this observation, we propose \textbf{HALMES}\footnote{HALMES is named after Sherlock Holmes, reflecting its role in detecting hallucination-prone token segments from internal attention patterns.}, a lightweight plug-and-play module that identifies hallucination precursor segments from attention features during generation and selectively triggers contrastive decoding. \textbf{HALMES} can be integrated with various full-stage contrastive decoding methods, transforming indiscriminate intervention into on-demand targeted correction. Extensive experiments across different LVLM architectures and multiple benchmarks show that \textbf{HALMES} further improves the hallucination mitigation performance of existing contrastive decoding methods while reducing the negative impact of full-stage intervention on normal outputs, validating the importance of modeling and selectively intervening on hallucination-prone segments.
\end{abstract}

\section{Introduction}
\label{sec:introduction}
Recent advances in Large Vision-Language Models (LVLMs) have demonstrated impressive capabilities across a wide range of multimodal tasks, including image captioning \cite{liu2023visual, liu2024improved, chen2024sharegpt4v, dai2023instructblip, dong2024benchmarking, cheng2025caparenabenchmarkinganalyzingdetailed, rohrbach2018object, fu2023mme, liu2024mmbench}, visual question answering \cite{antol2015vqa, goyal2017making, hudson2019gqa, gurari2018vizwiz, marino2019ok}, and visual reasoning \cite{krishna2017visual, johnson2017clevr, lu2022learn, schwenk2022okvqa}. However, the ever-growing adoption of LVLMs also gives rise to concerns about the reliability of their outputs. In particular, research has indicated that LVLMs are susceptible to object hallucination, generating textual descriptions that contain objects inconsistent with the given visual input \cite{ji2023survey, huang2025survey, rawte2023survey, bai2024hallucination, li2023evaluating, sun2023aligninglargemultimodalmodels}. This tendency not only undermines the trustworthiness of model outputs but also poses potential safety risks in high-stakes applications. Consequently, the reliable detection and effective mitigation of hallucinations are critical for establishing trustworthy LVLMs.

Various approaches have been developed to mitigate object hallucination in LVLMs. Among them, contrastive decoding has attracted considerable attention as a training-free inference-time intervention paradigm. The core idea is to construct a biased contrastive branch relative to the original inference process and suppress hallucination-prone tokens by contrasting the output distributions of the two branches. Depending on how the contrastive signal is constructed, existing methods can be broadly categorized into two lines: one line perturbs the visual or textual input to alleviate visual attention drift \cite{leng2024mitigating, chen2024halc, favero2024multi, liu2024paying, im2026selfaugqueryentropyadaptive, woo2025dontmissforesttrees} and suppress inherent biases from the language modality \cite{wang2024mitigating, li2023contrastivedecodingopenendedtext}, while another line contrasts output distributions across different layers, leveraging inter-layer information discrepancies to identify and correct hallucinated content \cite{chuang2023dola, wang2024valid}. Despite their diverse strategies, these methods share a common paradigm: intervention is applied to the token-level output probability distribution at every decoding step. However, we observe that this indiscriminate, full-stage intervention strategy introduces non-negligible side effects, as it can degrade originally correct outputs and even induce hallucinations in samples that would be hallucination-free without intervention. This observation suggests that contrastive decoding should not be applied in a one-size-fits-all manner. Instead, intervention should be selectively triggered only at positions where hallucination risk is genuinely elevated. This necessitates a mechanism capable of explicitly localizing the onset of hallucination during generation, thereby enabling targeted intervention.

To enable targeted intervention, it is essential to detect hallucination at the token level during generation. Existing token-level hallucination detection methods typically focus on identifying hallucinated object nouns, determining whether a given noun token refers to a real or hallucinated object \cite{fieback2025efficient, niu2025robust, prabhakaran2025vade, jiang2025devils, fieback2024metatoken}. However, recent studies provide compelling evidence that hallucination does not necessarily originate from the hallucinated noun token itself. Several works \cite{duan2025truthprint, huang2024opera, zhang2026hallucinationbeginssaliencydrops} have shown that hallucination-related triggers may emerge as continuous patterns in preceding tokens before the hallucinated object is generated. These findings reveal a critical gap: existing methods can distinguish hallucinated nouns from truthful ones in a post-hoc manner, but isolated token-level decisions are insufficient to capture the progressive anomalous patterns that lead to hallucination during real-time generation. To selectively trigger contrastive decoding, the detector should determine whether the current decoding trajectory has entered a hallucination-prone segment, rather than whether a single token position appears abnormal. We therefore formulate hallucination detection as a token-segment-level problem that captures anomalous patterns across the preceding token sequence.

Motivated by this challenge, we investigate cross-modal fusion in LVLMs by analyzing attention distributions between textual and visual modalities. Specifically, we adopt the Visual Attention Ratio (VAR) metric proposed in \cite{jiang2025devils}, which characterizes the degree of attention that each generated token allocates to image tokens across layers and attention heads. Meanwhile, inspired by \cite{fieback2024metatoken, chuang2024lookback}, we further design a set of complementary attention-based implicit features that capture the model's reliance on different modalities from multiple perspectives during generation. Through systematic statistical analysis, we find that hallucinated token segments and normal token segments exhibit significant distributional differences across all of these implicit features. To verify the applicability of these implicit features across different model architectures, we conduct experiments on two representative mainstream LVLMs \cite{bai2025qwen3, wang2025internvl3}. Although the specific anomaly distributions vary across models, the designed implicit features consistently capture hallucination-related anomalous patterns, demonstrating their effectiveness as architecture-agnostic hallucination indicators.

Based on the above empirical findings, we propose HALMES, a lightweight plug-and-play module that detects anomalous token segments during LVLM inference and selectively triggers contrastive decoding. In HALMES, we depart from the conventional token-level noun classification paradigm and instead reformulate hallucination detection as a token-segment-level binary classification problem, capturing hallucination precursor patterns from the attention features of the generated sequence. As an external module, HALMES can be seamlessly deployed on top of any full-stage contrastive decoding method, transforming indiscriminate intervention into on-demand, targeted correction. Extensive experimental results demonstrate that the integration of HALMES further improves the hallucination mitigation performance of full-stage contrastive decoding, which also highlights the importance of modeling and intervening on hallucination-prone anomalous segments.

In summary, our contributions are as follows: 
\begin{itemize}
    \item We systematically analyze the relationship between internal attention features and hallucinated token segments in LVLMs, revealing the high specificity of attention-based features as hallucination indicators and verifying their generality across different LVLM architectures.
    \item To the best of our knowledge, we are the first to propose a token-segment-level hallucination detection method, which extends hallucination detection from post-hoc classification of individual noun tokens to real-time anomaly identification over token segments during generation. The proposed method HALMES detects hallucination precursors from anomalous patterns in the attention feature space and integrates with multiple contrastive decoding methods in a plug-and-play manner, transforming full-stage indiscriminate intervention into on-demand targeted intervention.
    \item We conduct comprehensive experiments with multiple contrastive decoding methods across several widely-used benchmarks. The results validate the efficacy of token-segment-level hallucination detection and targeted intervention, and demonstrate that HALMES significantly reduces the side effects of existing state-of-the-art contrastive decoding methods, yielding further improvements in hallucination mitigation performance.
\end{itemize}

\section{Related Work}
\textbf{Internal Representations for Hallucination Detection.  } Recent studies have demonstrated that the internal representations of large models contain discriminative signals indicative of hallucination, and have been widely adopted for hallucination detection. At the response level, several methods \cite{chen2024inside, kossen2024semantic, manakul2023selfcheckgpt} assess the overall trustworthiness of generated responses through the semantic consistency of hidden states or the uncertainty of output distributions, yet lack the ability to pinpoint where hallucinations occur within the sequence. This limitation has motivated a shift toward token-level detection. In the LLM domain, works such as \cite{chuang2024lookback, zhang2025icr, niu2025robust, binkowski2026attention} leverage attention map features, cross-layer hidden state dynamics, or adaptive token selection to identify hallucinated tokens. In the LVLM domain, several approaches \cite{jiang2025devils, duan2025truthprint, fieback2024metatoken, che2025hallucinatory, fieback2025efficient} further incorporate visual attention patterns or cross-modal hidden states to detect hallucinated object nouns. Despite their notable progress, these token-level methods remain confined to isolated token positions and cannot capture anomalous patterns that may emerge within the token segment preceding a hallucinated noun. Recent work \cite{ogasa2025hallucinated} extends detection to the span level but is limited to post-hoc labeling in LLM text generation without real-time triggering during inference. Overall, existing methods cannot yet identify hallucination precursors at the token-segment level during inference. In contrast, HALMES operates at the token-segment level, capturing anomalous attention patterns preceding potential hallucinations to enable on-demand triggering of contrastive decoding.

\textbf{Contrastive Decoding for Hallucination Mitigation.  } Contrastive decoding suppresses hallucination components by constructing a contrastive branch that deviates from the original inference process, and has become a mainstream paradigm for hallucination mitigation. Existing methods either amplify hallucination biases by perturbing the visual or textual input and subtract them from the original distribution \cite{leng2024mitigating, favero2024multi, wang2024mitigating, manevich2024mitigating, woo2024ritual, zhu2025ibd, im2026selfaugqueryentropyadaptive}, or operate from the model's internals by contrasting output distributions across layers or adjusting intermediate attention weights to enhance visual information utilization \cite{chuang2023dola, liu2024paying, jiang2025devils, deng2025maskcd, woo2025dontmissforesttrees}. However, these methods all apply intervention at every decoding step, which may degrade originally correct outputs. To mitigate this side effect, recent works \cite{huang2024opera, zhang2026hallucinationbeginssaliencydrops, suo2025octopus} have explored targeted intervention strategies that detect anomalous patterns or adaptively select intervention schemes, achieving improved performance. Nevertheless, their triggering mechanisms still rely on heuristic rules or single-step feature judgments, lacking systematic modeling of the hallucination generation process. To this end, HALMES leverages the token-segment-level detection described above to enable plug-and-play, on-demand triggering of contrastive decoding.


\section{Preliminaries}
\textbf{Autoregressive Generation in LVLMs.  } Given an LVLM $\mathcal{M}_\theta$ parameterized by $\theta$, the input consists of an image $\mathbf{I}\in \mathbb{R}^{H \times W \times 3}$ and a text query $x$. The text query is tokenized into a sequence of tokens $\{\mathbf{x}_1, \dots, \mathbf{x}_m\}$, and a vision encoder maps $\mathbf{I}$ into a set of $d$-dimensional visual tokens $\{\mathbf{v}_1, \dots, \mathbf{v}_n\}$, which are then projected into the shared embedding space of the language model via a modality alignment module. Subsequently, the model generates a response in an autoregressive manner: at timestep $t$, the model processes the input tokens $\{\mathbf{v}_1, \dots, \mathbf{v}_n, \mathbf{x}_1, \dots, \mathbf{x}_m\}$, followed by the previously generated tokens $\{\mathbf{y}_1, \dots, \mathbf{y}_{t-1}\}$, to predict the next token $\mathbf{y}_t$. Let $y_t \in \mathcal{V}$ denote the index of $\mathbf{y}_t$ in the vocabulary $\mathcal{V}$. The probability distribution over $\mathcal{V}$ is given by:
\begin{equation}
p_\theta(y_t \mid \mathbf{I}, x, y_{<t}) = \mathrm{softmax}(\mathrm{logit}_\theta(y_t \mid \mathbf{I}, x, y_{<t})), 
\end{equation}
where $\mathrm{logit}_\theta(\cdot)$ denotes the unnormalized scores from the final layer. The next token is then sampled from this distribution according to the chosen decoding strategy.

\textbf{Contrastive Decoding.  } Contrastive decoding mitigates hallucinations by subtracting hallucination-amplified logits from the original logits before the softmax operation. Specifically, given the original input $(\mathbf{I},x)$, a contrastive branch is constructed by applying a predefined perturbation to the visual input $\mathbf{I}$, the text query $x$, or both, yielding a distorted input that amplifies hallucination biases. The contrastive distribution is computed as:
\begin{equation}
p_{\mathrm{cd}}(y_t \mid \mathbf{I}, \mathbf{I'}, x, x', y_{<t}) = \mathrm{softmax}\big( 
(1 + \alpha) \cdot \mathrm{logit}_\theta(y_t \mid \mathbf{I}, x, y_{<t})
- \alpha \cdot \mathrm{logit}_\theta(y_t \mid \mathbf{I'}, x', y_{<t})\big),
\end{equation}
where $\alpha$ controls the strength of hallucination suppression, with $\alpha = 0$ reducing to standard decoding. Various sampling strategies \cite{freitag2017beam, holtzman2019curious} can be applied to $p_{\mathrm{cd}}$. Notably, this correction is applied at every timestep $t = 1, 2, \dots, T$, uniformly modifying the output distribution regardless of whether the current token is at risk of hallucination. As discussed in Section~\ref{sec:introduction}, such indiscriminate full-stage intervention may unnecessarily perturb correct outputs and even introduce new hallucinations.

\section{Methodology}
\subsection{Attention-based Hallucination Pattern Analysis}
\textbf{Why Attention-based Features? } Recent studies have shown that the internal states of LVLMs contain discriminative signals for hallucination detection. Existing approaches have explored such signals from different perspectives, including hidden-state representations \cite{duan2025truthprint}, generation statistics \cite{fieback2024metatoken}, and attention patterns \cite{jiang2025devils}. These findings suggest that object hallucination is not only reflected in the final output, but also leaves observable traces in the model’s internal dynamics. In LVLMs, object hallucination is often linked to excessive reliance on language priors and insufficient visual grounding during cross-modal fusion, the latter typically manifested as visual attention drift or imbalanced dependence across modalities. For the hallucination precursor segments studied in this work, hidden states are less explicit in revealing how the model reallocates dependence across different information sources, while generation statistics mainly characterize output behavior rather than the internal fusion process itself. By contrast, attention distributions directly capture how the current decoding position attends to visual tokens, prompt tokens, and previously generated text. Therefore, attention-based features provide a more suitable basis for our analysis and modeling.

\textbf{Features Formulation. } Based on the above analysis, we construct a set of implicit features derived from cross-modal attention distributions to characterize the internal state of the current decoding position in terms of visual reliance, textual reliance, the relative balance between textual and visual reliance, and the complexity of the outgoing attention distribution.
Let $\mathbf{y}_t$ denote the token generated at decoding step $t$. For the $h$-th attention head in the $l$-th layer, we denote by:
\begin{equation}
    \mathbf{a}_t^{(l,h)} \in \mathbb{R}^{N_t}, 
\end{equation}
as the attention distribution of the current decoding position over all visible tokens, where $N_t$ is the number of visible tokens at step $t$. We further denote by $\mathcal{V}$, $\mathcal{P}$, and $\mathcal{G}$ the sets of visual tokens, prompt tokens, and previously generated text tokens, respectively. Under these notations, we define the attention-based features used in our method as follows.

\begin{itemize}
    \item \emph{Visual attention ratio (VAR).} Following \cite{jiang2025devils}, we use VAR to quantify the attention allocated to visual tokens:
    \begin{equation}
    \mathrm{VAR}_t^{(l,h)}=\sum_{i\in \mathcal{V}} a_{t,i}^{(l,h)}, \qquad
    \mathrm{SVAR}_t=\sum_{l=1}^{L}\left(\frac{1}{H}\sum_{h=1}^{H}\mathrm{VAR}_t^{(l,h)}\right),
    \end{equation}
    where $\mathrm{SVAR}_t$ provides a compact measure of the overall visual reliance at decoding step $t$.

    \item \emph{Prompt attention.} To measure the dependence on the input prompt, we define:
    \begin{equation}
    A_t^{\mathrm{prompt}}
    =
    \frac{1}{L}\sum_{l=1}^{L}
    \left(
    \frac{1}{H|\mathcal{P}|}
    \sum_{h=1}^{H}\sum_{i\in\mathcal{P}} a_{t,i}^{(l,h)}
    \right),
    \end{equation}
    which reflects the extent to which the current decoding position attends to the prompt tokens.

    \item \emph{Generated-text attention.} We define the average attention to prior generated tokens as:
    \begin{equation}
    A_t^{\mathrm{gen}}
    =
    \frac{1}{L}\sum_{l=1}^{L}
    \left(
    \frac{1}{H|\mathcal{G}|}\sum_{h=1}^{H}\sum_{i\in\mathcal{G}} a_{t,i}^{(l,h)}
    \right),
    \end{equation}
    with this quantity set to zero when no previously generated text is available.

    \item \emph{Text-visual ratio.} To quantify the balance between textual and visual reliance, we define:
    \begin{equation}
    R_t^{\mathrm{text/vis}}
    =
    \frac{A_t^{\mathrm{gen}}}{\mathrm{SVAR}_t/L},
    \end{equation}
    where larger values indicate higher generated-text attention relative to visual attention.

    \item \emph{Outgoing attention entropy.} To quantify the dispersion of the current token's attention over all visible tokens, we define the head-level entropy as:
    \begin{equation}
    \mathcal{H}_t^{(l,h)}
    =
    -\sum_{i=1}^{N_t} a_{t,i}^{(l,h)}
    \log a_{t,i}^{(l,h)},
    \end{equation}
    with the final normalized outgoing attention entropy given by:
    \begin{equation}
    E_t
    =
    \frac{1}{L}\sum_{l=1}^{L}
    \left(
    \frac{1}{H}\sum_{h=1}^{H}
    \frac{\mathcal{H}_t^{(l,h)}}{\log N_t}
    \right),
    \end{equation}
    where $E_t$ measures the degree of concentration of attention at the current decoding position.
\end{itemize}


\begin{figure*}[t]
    \centering
    \includegraphics[width=\textwidth]{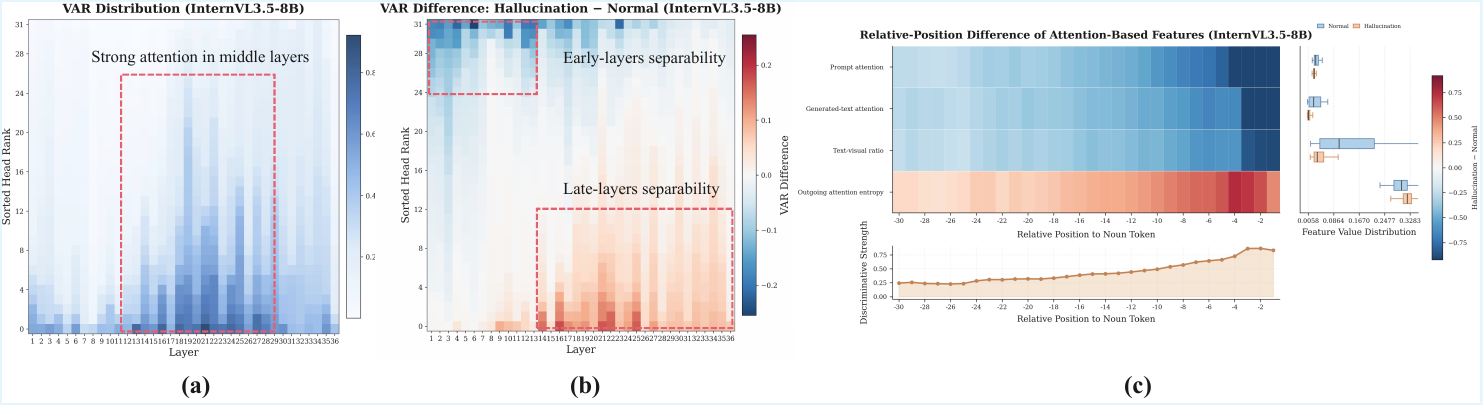}
    \vspace{-4mm}
    \caption{Attention-based feature analysis for hallucinated and normal token segments. (a) Average VAR distribution across layers and attention heads. (b) Layer-wise VAR difference between hallucinated and normal samples. (c) Relative-position differences of prompt attention, generated-text attention, text-visual ratio, and outgoing attention entropy before the noun token.}
    \label{fig:feature_analysis}
    \vspace{-3mm}
\end{figure*}

\textbf{Hallucination Pattern Discovery.  } We construct a token-segment-level dataset from a randomly sampled subset of 5,000 images from MSCOCO train2014 \cite{lin2014microsoft}. For each image, we prompt the target LVLM to generate a response and use CHAIR to localize hallucinated object nouns. Based on the localized noun positions, each response is decomposed into multiple token-segment samples. Specifically, for each noun token, we collect the attention-based features of all tokens from the beginning of the response to the position immediately before that noun token, and assign the segment a hallucinated or normal label according to whether the corresponding noun is hallucinated. To avoid ambiguous supervision, we discard all non-hallucinated noun segments that appear after the first hallucinated noun in the same response, since the subsequent generation may already deviate from the normal decoding trajectory. As a result, a single response may yield multiple labeled token-segment samples for the subsequent statistical analysis.

To analyze how generated tokens attend to image tokens, we visualize the average VAR distribution across layers and heads in Fig.~\ref{fig:feature_analysis}(a). Visual attention is mainly concentrated in the middle layers and a few highly responsive heads, which is consistent with prior observations in~\cite{jiang2025devils} and suggests active visual information interaction in the middle layers. Fig.~\ref{fig:feature_analysis}(b) further shows the average VAR difference between hallucinated and normal samples. Normal samples exhibit stronger visual attention in early layers, indicating earlier visual grounding, whereas hallucinated samples show higher visual attention in middle and deep layers, with the difference gradually weakening toward deeper layers. This suggests that later increases in visual attention do not necessarily correct an already deviated generation trajectory. Overall, VAR across the network carries hallucination-related discriminative information, motivating us to retain the full-layer VAR representation in subsequent modeling.

For the remaining four features, we analyze hallucinated--normal differences at positions preceding the noun token, as shown in Fig.~\ref{fig:feature_analysis}(c). As decoding approaches the noun token, prompt attention, generated-text attention, and text-visual ratio consistently show negative differences, indicating weaker reliance on the prompt and generated history in hallucinated samples before the noun is produced. In contrast, outgoing attention entropy shows positive and increasingly pronounced differences, suggesting more diffuse attention and weaker focus in hallucinated samples. The right-side distributions further show stable separability between the two groups, while the bottom discriminative-strength curve indicates that such differences become stronger near the noun token. These results support our hypothesis that hallucination precursors emerge progressively within the preceding token segment rather than only at the noun position.

\subsection{HALMES}
\textbf{Token-Segment-Level Detection Model.  }
During training, each sample is represented as a prefix feature sequence and passed to a lightweight gate model $g_{\phi}$ for hallucination precursor prediction. Let $\mathbf{z}_t \in \mathbb{R}^{D}$ denote the feature vector at decoding step $t$, where $\mathbf{z}_t$ is composed of the attention-based features defined above. For a training sample, let $n-1$ denote the length of the generated prefix preceding its corresponding noun token. The input sequence is then defined as:
\begin{equation}
\mathbf{Z}_n=
[\mathbf{z}_{\,n-\lfloor W(n-1)\rfloor},\mathbf{z}_{\,n-\lfloor W(n-1)\rfloor+1},\dots,\mathbf{z}_{\,n-1}]
\in\mathbb{R}^{\lfloor W(n-1)\rfloor\times D},
\end{equation}
where $W\in(0,1]$ denotes the retained tail ratio. Therefore, $\mathbf{Z}_n$ is a local token-level feature segment that captures the attention-pattern evolution before the corresponding noun token. We aggregate $\mathbf{Z}_n$ along the token dimension into a fixed-dimensional segment representation $\mathbf{s}_n$, which is then fed into the gate model to output a hallucination risk score:
\begin{equation}
p_n=\mathrm{sigmoid}\big(g_{\phi}(\mathbf{s}_n)\big)\in[0,1].
\end{equation}
In our implementation, $g_{\phi}$ is instantiated as a 2-layer MLP backbone, offering a lightweight and effective architecture that is well suited for efficient deployment at inference time.

\textbf{Inference-time Deployment.  }
During inference, we attach the gate as an external lightweight module to the original decoding process. Let $\mathbf{z}_t \in \mathbb{R}^{D}$ denote the feature vector extracted at decoding step $t$, which is appended to a feature buffer. To reduce distribution shifts across responses, we use the training-set mean and variance as priors and perform one Welford update \cite{welford1962note} with the first sample of the current response. The resulting statistics are then fixed and used to normalize all subsequent window features for the rest of the response. At each step $t$, we extract the most recent $K$ feature vectors from the buffer to form a local window and compute:
\begin{equation}
\hat{\mathbf Z}_t
=
\left[
\frac{\mathbf z_{t-K+1}-\boldsymbol\mu}{\boldsymbol\sigma},
\dots,
\frac{\mathbf z_t-\boldsymbol\mu}{\boldsymbol\sigma}
\right],
\qquad
\mathbf{s}_t=\mathrm{Agg}(\hat{\mathbf Z}_t),
\qquad
p_t=\mathrm{sigmoid}\big(g_\phi(\mathbf{s}_t)\big),
\end{equation}
where $\boldsymbol\mu$ and $\boldsymbol\sigma$ denote the mean and standard deviation after prior initialization and one-step adaptive calibration, $K$ is the sliding-window size, $\mathrm{Agg}(\cdot)$ is the training-time aggregation, and $p_t\in[0,1]$ is the hallucination risk score. To remain consistent with the segment-level training paradigm, we apply the gate to aggregated local token segments rather than isolated token-level features. In practice, $K$ is set slightly smaller than the average length of the training samples so that the input segment remains close to that used in training while still enabling early warning of abnormal segments. We further apply a trend-aware double-threshold hysteresis mechanism to the score sequence to reduce false triggering caused by local noise. The detailed inference procedure is provided in the appendix.

\section{Experiments}
We evaluated the efficacy of HALMES through extensive experiments on multiple benchmarks. Specifically, we focused on the following research questions.
\begin{itemize}
    \item[\textbf{Q1.}] Whether the proposed attention-based features can support token-segment-level hallucination detection across different LVLM architectures.
    \item[\textbf{Q2.}] Whether selectively applying contrastive decoding to detected anomalous token segments can preserve hallucination mitigation efficacy while reducing the side effects of indiscriminate full-stage intervention.
\end{itemize}
Further details on the implementation and additional experimental results are provided in the appendix.

\subsection{Q1: Attention-based Features for Token-Segment-Level Hallucination Detection}
To validate whether the proposed attention-based features can support hallucination-prone segment detection, we varied the VAR layer range while keeping the remaining four features fixed to examine the layer-wise distribution of hallucination-related cues.

The results are shown in Table~\ref{tab:hallucination_detection}. On both LVLM architectures, full-layer VAR achieves the highest AUROC, indicating that the proposed attention-based features can serve as useful indicators for token-segment-level hallucination detection. Meanwhile, the second-best layer range differs across architectures: middle-layer VAR (13--24) performs second-best on Qwen3VL-8B, whereas deep-layer VAR (25--36) is more competitive on InternVL3.5-8B. This suggests that hallucination-related cues are distributed across layers, while their most prominent ranges vary across LVLM architectures. Considering that full-stage contrastive decoding may introduce side effects on normal outputs, we paid particular attention to whether the detector could maintain high Recall while controlling the false positive rate (FPR). Although full-layer VAR does not always achieve the lowest FPR, it obtains the highest Recall on both models and provides a better trade-off with second-best or near-second-best FPR. Therefore, we adopted full-layer VAR as the input for subsequent modeling.



\begin{table*}[t]
\centering
\small
\setlength{\tabcolsep}{3.2pt}
\renewcommand{\arraystretch}{1.12}
\caption{Hallucination detection performance under different VAR layer ranges on two LVLM architectures \cite{bai2025qwen3, wang2025internvl3}. All settings include the remaining four features and differ only in the chosen VAR layer range. The \colorbox{bestcolor}{\textbf{best}} and \colorbox{secondcolor}{second-best} results are highlighted in color.}
\label{tab:hallucination_detection}
\vspace{2pt}
\resizebox{\textwidth}{!}{%
\begin{tabular}{l|ccccc|ccccc}
\toprule
\multirow{2}{*}[-0.5ex]{VAR layers} & \multicolumn{5}{c|}{\textbf{Qwen3VL-8B}} & \multicolumn{5}{c}{\textbf{InternVL3.5-8B}} \\
\cmidrule(lr){2-6} \cmidrule(lr){7-11}
& AUROC {\scriptsize$\uparrow$} & Accuracy {\scriptsize$\uparrow$} & Recall {\scriptsize$\uparrow$} & PR-AUC {\scriptsize$\uparrow$} & FPR {\scriptsize$\downarrow$}
& AUROC {\scriptsize$\uparrow$} & Accuracy {\scriptsize$\uparrow$} & Recall {\scriptsize$\uparrow$} & PR-AUC {\scriptsize$\uparrow$} & FPR {\scriptsize$\downarrow$} \\
\midrule
1--12  
& 88.35 & 80.00 & 72.89 & 88.72 & 18.22
& 88.28 & 79.61 & 82.14 & 88.05 & 22.92 \\

13--24 
& \colorbox{secondcolor}{90.64} & \colorbox{secondcolor}{84.00} & \colorbox{secondcolor}{84.89} & \colorbox{bestcolor}{\textbf{90.82}} & 16.89
& 88.92 & 81.40 & 81.85 & 87.57 & \colorbox{bestcolor}{\textbf{19.06}} \\

25--36 
& 89.91 & 83.11 & 76.89 & \colorbox{secondcolor}{90.26} & \colorbox{bestcolor}{\textbf{10.67}}
& \colorbox{secondcolor}{90.24} & \colorbox{secondcolor}{82.59} & \colorbox{secondcolor}{86.01} & \colorbox{secondcolor}{88.78} & 20.83 \\

\midrule
\textbf{1--36}
& \colorbox{bestcolor}{\textbf{91.31}} & \colorbox{bestcolor}{\textbf{86.73}} & \colorbox{bestcolor}{\textbf{86.39}} & 88.76 & \colorbox{secondcolor}{12.93}
& \colorbox{bestcolor}{\textbf{90.52}} & \colorbox{bestcolor}{\textbf{83.63}} & \colorbox{bestcolor}{\textbf{87.20}} & \colorbox{bestcolor}{\textbf{90.03}} & \colorbox{secondcolor}{19.94} \\
\bottomrule
\end{tabular}%
}
\end{table*}

\begin{figure*}[t]
    \centering
    \includegraphics[width=\textwidth]{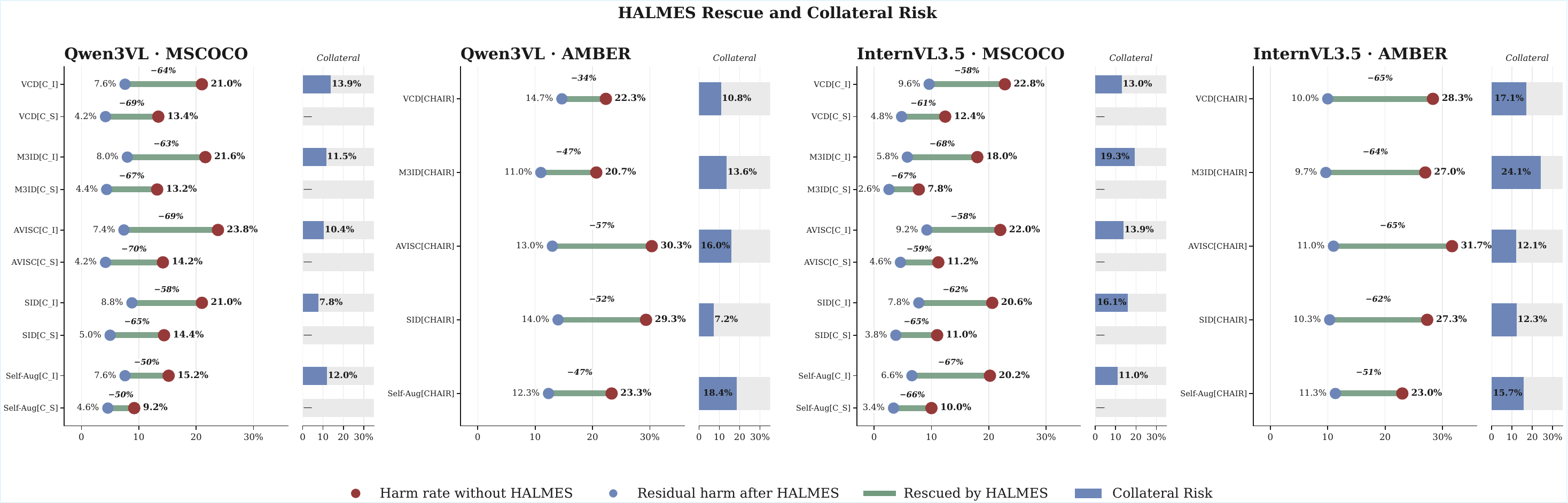}
    \vspace{-4mm}
    \caption{Sample-level rescue and collateral risk analysis of HALMES. HALMES reduces harmful cases caused by full-stage contrastive decoding, while the right-side bars show the collateral risk introduced by selective triggering.}
    \label{fig:halmes_analysis}
    \vspace{-4mm}
\end{figure*}

\subsection{Q2: Selective Contrastive Decoding on Detected Anomalous Token Segments}
\textbf{Object Hallucination Mitigation on MSCOCO and AMBER.  }
As shown in Table~\ref{tab:coco_amber}, we evaluated the effect of HALMES on object hallucination mitigation using MSCOCO val2014 \cite{lin2014microsoft} and AMBER \cite{wang2023amber}. The results show that integrating HALMES into five full-stage contrastive decoding methods further reduces hallucination metrics in most settings, indicating that contrastive decoding does not necessarily need to be applied indiscriminately throughout the generation process. Instead, restricting intervention to detected anomalous token segments can preserve hallucination mitigation efficacy while potentially improving it. For MSCOCO, different contrastive decoding methods exhibit different behaviors: some methods reduce hallucination, whereas others even aggravate hallucination under full-stage intervention. After introducing HALMES, all CD methods obtain consistent improvements, suggesting that selective triggering can alleviate the instability of full-stage intervention. Meanwhile, the visual ground-truth coverage (Cov.) decreases for some methods, indicating that HALMES may make descriptions slightly more conservative while reducing hallucinated object mentions. This reflects a potential trade-off between hallucination mitigation and visual coverage. For AMBER, HALMES reduces hallucination in most settings, with a maximum HalRate reduction of 27.6\%, though its gains are limited for M3ID and Self-Aug. This indicates that selective triggering can behave differently across CD methods and benchmark settings. Overall, these results support our central claim that selective contrastive decoding over anomalous token segments can mitigate hallucination more effectively than indiscriminate full-stage intervention in most settings.


\begin{table*}[t]
\centering
\caption{Hallucination evaluation results on MSCOCO and AMBER for Qwen~\cite{bai2025qwen3} and InternVL~\cite{wang2025internvl3}. Arrows indicate the preferred direction. \textbf{Bold} denotes the best result. In the $\Delta$ blocks, {\color{improvecolor}+} and {\color{degradecolor}-} indicate improvement and degradation over the corresponding contrastive decoding method.}
\label{tab:coco_amber}
\vspace{2pt}
\resizebox{\textwidth}{!}{%
\begin{tabular}{l|ccc|cccc|ccc|cccc}
\toprule
\multirow{3}{*}{\textbf{Method}} & \multicolumn{7}{c|}{\textbf{Qwen3VL-8B}} & \multicolumn{7}{c}{\textbf{InternVL3.5-8B}} \\
\cmidrule(lr){2-8}\cmidrule(lr){9-15}
& \multicolumn{3}{c|}{\textbf{MSCOCO}} & \multicolumn{4}{c|}{\textbf{AMBER}}
& \multicolumn{3}{c|}{\textbf{MSCOCO}} & \multicolumn{4}{c}{\textbf{AMBER}} \\
\cmidrule(lr){2-4}\cmidrule(lr){5-8}\cmidrule(lr){9-11}\cmidrule(lr){12-15}
& C$_I$$\downarrow$ & C$_S$$\downarrow$ & Cov.$\uparrow$& CHAIR$\downarrow$ & Cover$\uparrow$ & Cog$\downarrow$ & Hal$\downarrow$& C$_I$$\downarrow$ & C$_S$$\downarrow$ & Cov.$\uparrow$& CHAIR$\downarrow$ & Cover$\uparrow$ & Cog$\downarrow$ & Hal$\downarrow$ \\
\midrule
\rowcolor{groupcolorA}\multicolumn{15}{l}{\textit{\textbf{Base}}} \\
Beam   & 13.35 & 30.80 & 52.26 & 7.48 & 53.55 & 3.52 & 35.33 & 18.02 & \textbf{30.00} & 37.20 & 11.73 & 43.22 & 3.67 & 41.00 \\
Nucleus& 13.84 & 33.60 & 51.20 & 8.82 & 52.89 & 3.02 & 35.60 & 16.07 & 43.20 & 66.94 & 6.37 & 52.46 & 3.72 & 34.33 \\
\midrule
\rowcolor{groupcolorA}\multicolumn{15}{l}{\textit{\textbf{Full-Stage Contrastive Decoding}}} \\
VCD {\footnotesize\textit{(CVPR 2024)}}    & 12.96 & 35.40 & 51.96 & 7.56 & 52.25 & 3.65 & 37.33 & 14.92 & 44.00 & 68.22 & 7.34 & \textbf{59.20} & 4.06 & 38.67 \\
M3ID {\footnotesize\textit{(CVPR 2024)}}    & 14.50 & 35.80 & 50.98 & 7.37 & 52.99 & 3.03 & \textbf{34.67} & 13.26 & 37.80 & 67.92 & 7.25 & 54.71 & 3.16 & 38.00 \\
AVISC {\footnotesize\textit{(ACL 2025)}}  & 14.26 & 38.80 & 53.39 & 8.86 & 53.26 & 3.53 & 43.00 & 14.87 & 43.60 & 69.35 & 7.75 & 55.04 & 3.21 & 41.00 \\
SID {\footnotesize\textit{(ICLR 2025)}}   & 12.74 & 37.40 & 53.39 & 9.12 & 52.91 & 3.53 & 40.33 & 14.07 & 41.40 & 69.50 & 7.55 & 54.94 & 3.24 & 37.33 \\
Self-Aug {\footnotesize\textit{(ICLR 2026)}} & 12.11 & 31.20 & 51.66 & \textbf{7.33} & 52.16 & \textbf{3.01} & 35.33 & 13.92 & 40.80 & 68.30 & 6.14 & 54.58 & 2.98 & 32.33 \\
\midrule
\rowcolor{groupcolorB}\multicolumn{15}{l}{\textit{\textbf{HALMES (Ours)}}} \\
\rowcolor{groupcolorB!30}
HALMES+VCD   & 11.50 & 31.80 & \textbf{54.14} & 7.52 & 54.18 & 3.23 & 37.33 & 13.00 & 38.20 & 68.30 & 6.18 & 54.93 & 2.98 & \textbf{28.00} \\
\rowcolor{groupcolorB!30}
HALMES+M3ID  & \textbf{10.76} & 30.00 & 53.69 & 7.67 & 52.57 & 3.56 & 37.33 & 12.01 & 35.60 & 68.00 & 7.76 & 54.79 & 3.91 & 34.67 \\
\rowcolor{groupcolorB!30}
HALMES+AVISC & 11.41 & \textbf{29.00} & 52.03 & 8.04 & 54.42 & 3.40 & 39.67 & \textbf{12.00} & 38.40 & 69.05 & 7.46 & 54.27 & 3.41 & 33.33 \\
\rowcolor{groupcolorB!30}
HALMES+SID   & 11.44 & 31.80 & 53.24 & 8.19 & \textbf{54.66} & 3.45 & 40.00 & 12.30 & 37.80 & 67.62 & \textbf{5.98} & 55.34 & \textbf{2.89} & 32.67 \\
\rowcolor{groupcolorB!30}
HALMES+Self-Aug & 11.27 & 30.60 & 53.24 & 7.84 & 53.79 & 3.08 & 38.33 & 12.58 & 36.40 & \textbf{70.11} & 6.52 & 54.05 & 3.16 & 30.33 \\
\midrule
\rowcolor{groupcolorB}\multicolumn{15}{l}{\textit{\textbf{$\Delta$ absolute (pp) vs. corresponding contrastive decoding method} \; ({\color{improvecolor}+}: improvement, {\color{degradecolor}-}: degradation)}} \\
\rowcolor{groupcolorB!30}
HALMES+VCD
 & {\color{improvecolor}+1.46} & {\color{improvecolor}+3.60} & {\color{improvecolor}+2.18}
 & {\color{improvecolor}+0.04} & {\color{improvecolor}+1.93} & {\color{improvecolor}+0.42} & {\color{gray}\;\;0.00}
 & {\color{improvecolor}+1.92} & {\color{improvecolor}+5.80} & {\color{improvecolor}+0.08}
 & {\color{improvecolor}+1.16} & {\color{degradecolor}-4.27} & {\color{improvecolor}+1.08} & {\color{improvecolor}+10.67} \\
\rowcolor{groupcolorB!30}
HALMES+M3ID
 & {\color{improvecolor}+3.74} & {\color{improvecolor}+5.80} & {\color{improvecolor}+2.71}
 & {\color{degradecolor}-0.30} & {\color{degradecolor}-0.42} & {\color{degradecolor}-0.53} & {\color{degradecolor}-2.66}
 & {\color{improvecolor}+1.25} & {\color{improvecolor}+2.20} & {\color{improvecolor}+0.08}
 & {\color{degradecolor}-0.51} & {\color{improvecolor}+0.08} & {\color{degradecolor}-0.75} & {\color{improvecolor}+3.33} \\
\rowcolor{groupcolorB!30}
HALMES+AVISC
 & {\color{improvecolor}+2.85} & {\color{improvecolor}+9.80} & {\color{degradecolor}-1.36}
 & {\color{improvecolor}+0.82} & {\color{improvecolor}+1.16} & {\color{improvecolor}+0.13} & {\color{improvecolor}+3.33}
 & {\color{improvecolor}+2.87} & {\color{improvecolor}+5.20} & {\color{degradecolor}-0.30}
 & {\color{improvecolor}+0.29} & {\color{degradecolor}-0.77} & {\color{degradecolor}-0.20} & {\color{improvecolor}+7.67} \\
\rowcolor{groupcolorB!30}
HALMES+SID
 & {\color{improvecolor}+1.30} & {\color{improvecolor}+5.60} & {\color{degradecolor}-0.15}
 & {\color{improvecolor}+0.93} & {\color{improvecolor}+1.75} & {\color{improvecolor}+0.08} & {\color{improvecolor}+0.33}
 & {\color{improvecolor}+1.77} & {\color{improvecolor}+3.60} & {\color{degradecolor}-1.88}
 & {\color{improvecolor}+1.57} & {\color{improvecolor}+0.40} & {\color{improvecolor}+0.35} & {\color{improvecolor}+4.66} \\
\rowcolor{groupcolorB!30}
HALMES+Self-Aug
 & {\color{improvecolor}+0.84} & {\color{improvecolor}+0.60} & {\color{improvecolor}+1.58}
 & {\color{degradecolor}-0.51} & {\color{improvecolor}+1.63} & {\color{degradecolor}-0.07} & {\color{degradecolor}-3.00}
 & {\color{improvecolor}+1.34} & {\color{improvecolor}+4.40} & {\color{improvecolor}+1.81}
 & {\color{degradecolor}-0.38} & {\color{degradecolor}-0.53} & {\color{degradecolor}-0.18} & {\color{improvecolor}+2.00} \\
\midrule
\rowcolor{groupcolorB}\multicolumn{15}{l}{\textit{\textbf{$\Delta$ relative (\%) vs. corresponding contrastive decoding method} \; ({\color{improvecolor}+}: improvement, {\color{degradecolor}-}: degradation)}} \\
\rowcolor{groupcolorB!30}
HALMES+VCD
 & {\color{improvecolor}+11.3\%} & {\color{improvecolor}+10.2\%} & {\color{improvecolor}+4.2\%}
 & {\color{improvecolor}+0.5\%} & {\color{improvecolor}+3.7\%} & {\color{improvecolor}+11.5\%} & {\color{gray}\;\;0.0\%}
 & {\color{improvecolor}+12.9\%} & {\color{improvecolor}+13.2\%} & {\color{improvecolor}+0.1\%}
 & {\color{improvecolor}+15.8\%} & {\color{degradecolor}-7.2\%} & {\color{improvecolor}+26.6\%} & {\color{improvecolor}+27.6\%} \\
\rowcolor{groupcolorB!30}
HALMES+M3ID
 & {\color{improvecolor}+25.8\%} & {\color{improvecolor}+16.2\%} & {\color{improvecolor}+5.3\%}
 & {\color{degradecolor}-4.1\%} & {\color{degradecolor}-0.8\%} & {\color{degradecolor}-17.5\%} & {\color{degradecolor}-7.7\%}
 & {\color{improvecolor}+9.4\%} & {\color{improvecolor}+5.8\%} & {\color{improvecolor}+0.1\%}
 & {\color{degradecolor}-7.0\%} & {\color{improvecolor}+0.1\%} & {\color{degradecolor}-23.7\%} & {\color{improvecolor}+8.8\%} \\
\rowcolor{groupcolorB!30}
HALMES+AVISC
 & {\color{improvecolor}+20.0\%} & {\color{improvecolor}+25.3\%} & {\color{degradecolor}-2.5\%}
 & {\color{improvecolor}+9.3\%} & {\color{improvecolor}+2.2\%} & {\color{improvecolor}+3.7\%} & {\color{improvecolor}+7.7\%}
 & {\color{improvecolor}+19.3\%} & {\color{improvecolor}+11.9\%} & {\color{degradecolor}-0.4\%}
 & {\color{improvecolor}+3.7\%} & {\color{degradecolor}-1.4\%} & {\color{degradecolor}-6.2\%} & {\color{improvecolor}+18.7\%} \\
\rowcolor{groupcolorB!30}
HALMES+SID
 & {\color{improvecolor}+10.2\%} & {\color{improvecolor}+15.0\%} & {\color{degradecolor}-0.3\%}
 & {\color{improvecolor}+10.2\%} & {\color{improvecolor}+3.3\%} & {\color{improvecolor}+2.3\%} & {\color{improvecolor}+0.8\%}
 & {\color{improvecolor}+12.6\%} & {\color{improvecolor}+8.7\%} & {\color{degradecolor}-2.7\%}
 & {\color{improvecolor}+20.8\%} & {\color{improvecolor}+0.7\%} & {\color{improvecolor}+10.8\%} & {\color{improvecolor}+12.5\%} \\
\rowcolor{groupcolorB!30}
HALMES+Self-Aug
 & {\color{improvecolor}+6.9\%} & {\color{improvecolor}+1.9\%} & {\color{improvecolor}+3.1\%}
 & {\color{degradecolor}-7.0\%} & {\color{improvecolor}+3.1\%} & {\color{degradecolor}-2.3\%} & {\color{degradecolor}-8.5\%}
 & {\color{improvecolor}+9.6\%} & {\color{improvecolor}+10.8\%} & {\color{improvecolor}+2.7\%}
 & {\color{degradecolor}-6.2\%} & {\color{degradecolor}-1.0\%} & {\color{degradecolor}-6.0\%} & {\color{improvecolor}+6.2\%} \\
\bottomrule
\end{tabular}%
}
\end{table*}

\textbf{General LVLM Evaluation on MMHal-Bench and Bingo.  }
We further evaluated HALMES on MMHal-Bench \cite{sun2023aligninglargemultimodalmodels} and Bingo \cite{cui2023holistic} to assess its impact on the model's generation capability, with results summarized in Table~\ref{tab:mmhal_bingo}. HALMES improves the hallucination rate and response score on MMHal-Bench in most settings, with relatively stable gains for VCD and AVISC, indicating that selective intervention still provides hallucination mitigation benefits in comprehensive evaluation scenarios. However, the improvements on Bingo are limited, and slight degradation is observed in some settings, suggesting that intervention on anomalous segments may affect response completeness or expression quality. Overall, HALMES has a positive effect on hallucination reduction, but in broader evaluation settings, it still requires a careful balance between hallucination mitigation and response quality.

\begin{table}[t]
\centering
\caption{Hallucination evaluation results on MMHal-Bench and Bingo for Qwen~\cite{bai2025qwen3} and InternVL~\cite{wang2025internvl3}. \textbf{Bold} denotes the best result. In the $\Delta$ block, {\color{improvecolor}+} and {\color{degradecolor}-} indicate improvement and degradation over the corresponding contrastive decoding method.}
\label{tab:mmhal_bingo}
\vspace{2pt}
\resizebox{0.6\textwidth}{!}{%
\begin{tabular}{l|ccc|ccc}
\toprule
\multirow{3}{*}{\textbf{Method}} & \multicolumn{3}{c|}{\textbf{Qwen3VL-8B}} & \multicolumn{3}{c}{\textbf{InternVL3.5-8B}} \\
\cmidrule(lr){2-4}\cmidrule(lr){5-7}
& \multicolumn{2}{c}{\textbf{MMHal-Bench}} & \textbf{Bingo}
& \multicolumn{2}{c}{\textbf{MMHal-Bench}} & \textbf{Bingo} \\
\cmidrule(lr){2-3}\cmidrule(lr){4-4}\cmidrule(lr){5-6}\cmidrule(lr){7-7}
& Hal$\downarrow$ & Score$\uparrow$ & Score$\uparrow$
& Hal$\downarrow$ & Score$\uparrow$ & Score$\uparrow$ \\
\midrule
\rowcolor{groupcolorA}\multicolumn{7}{l}{\textit{\textbf{Base}}} \\
Beam    & 45.26 & 3.81 & 3.54 & 63.83 & 2.36 & 2.43 \\
Nucleus & 47.37 & 3.71 & 3.52 & 37.89 & 3.66 & 3.18 \\
\midrule
\rowcolor{groupcolorA}\multicolumn{7}{l}{\textit{\textbf{Full-Stage Contrastive Decoding}}} \\
VCD {\footnotesize\textit{(CVPR 2024)}}     & 51.06 & 3.50 & 3.50 & 44.79 & 3.48 & 3.14 \\
M3ID {\footnotesize\textit{(CVPR 2024)}}    & 51.61 & 3.42 & 3.47 & 38.54 & 3.75 & 3.15 \\
AVISC {\footnotesize\textit{(ACL 2025)}}   & 44.68 & 3.69 & 3.43 & 47.92 & 3.38 & 3.09 \\
SID {\footnotesize\textit{(ICLR 2025)}}     & 43.01 & 3.83 & \textbf{3.56} & 42.55 & 3.59 & 3.24 \\
Self-Aug {\footnotesize\textit{(ICLR 2026)}} & 35.42 & 4.14 & 3.50 & 41.67 & 3.59 & 3.14 \\
\midrule
\rowcolor{groupcolorB}\multicolumn{7}{l}{\textit{\textbf{HALMES (Ours)}}} \\
\rowcolor{groupcolorB!30}
HALMES+VCD      & 40.66 & 3.79 & 3.49 & \textbf{36.84} & \textbf{3.76} & 3.14 \\
\rowcolor{groupcolorB!30}
HALMES+M3ID     & 49.46 & 3.57 & \textbf{3.56} & 41.05 & 3.74 & 3.12 \\
\rowcolor{groupcolorB!30}
HALMES+AVISC    & 42.55 & 3.87 & 3.37 & 42.71 & 3.68 & 3.10 \\
\rowcolor{groupcolorB!30}
HALMES+SID      & 48.35 & 3.36 & 3.54 & 41.67 & 3.68 & 3.13 \\
\rowcolor{groupcolorB!30}
HALMES+Self-Aug & \textbf{33.33} & \textbf{4.24} & 3.55 & 38.54 & 3.52 & \textbf{3.29} \\
\midrule
\rowcolor{groupcolorB}\multicolumn{7}{l}{\textit{\textbf{$\Delta$ vs. corresponding contrastive decoding method} \; ({\color{improvecolor}+}: improvement, {\color{degradecolor}-}: degradation)}} \\
\rowcolor{groupcolorB!30}
HALMES+VCD
 & {\color{improvecolor}+10.40} & {\color{improvecolor}+0.29} & {\color{degradecolor}-0.01}
 & {\color{improvecolor}+7.95}  & {\color{improvecolor}+0.28} & {\color{gray}\;\;0.00} \\
\rowcolor{groupcolorB!30}
HALMES+M3ID
 & {\color{improvecolor}+2.15} & {\color{improvecolor}+0.15} & {\color{improvecolor}+0.09}
 & {\color{degradecolor}-2.51}  & {\color{degradecolor}-0.01} & {\color{degradecolor}-0.03} \\
\rowcolor{groupcolorB!30}
HALMES+AVISC
 & {\color{improvecolor}+2.13} & {\color{improvecolor}+0.18} & {\color{degradecolor}-0.06}
 & {\color{improvecolor}+5.21}  & {\color{improvecolor}+0.30} & {\color{improvecolor}+0.01} \\
\rowcolor{groupcolorB!30}
HALMES+SID
 & {\color{degradecolor}-5.34} & {\color{degradecolor}-0.47} & {\color{degradecolor}-0.02}
 & {\color{improvecolor}+0.88}  & {\color{improvecolor}+0.09} & {\color{degradecolor}-0.11} \\
\rowcolor{groupcolorB!30}
HALMES+Self-Aug
 & {\color{improvecolor}+2.09} & {\color{improvecolor}+0.10} & {\color{improvecolor}+0.05}
 & {\color{improvecolor}+3.13}  & {\color{degradecolor}-0.07} & {\color{improvecolor}+0.15} \\
\bottomrule
\end{tabular}%
}
\end{table}

\textbf{Effect and Side-effect Analysis of HALMES.  }
To further characterize the benefits and potential side effects of HALMES, we analyzed the sample-level harm induced by full-stage CD and the corrective effect of HALMES, as shown in Fig.~\ref{fig:halmes_analysis}. Red points show cases where full-stage CD underperforms base decoding, blue points show the remaining harmful cases after HALMES, and green segments indicate rescued cases. Full-stage CD causes non-negligible degradation across models, datasets, and methods, whereas HALMES substantially reduces such harm and rescues around 60\% of degraded samples in most settings. The collateral-risk analysis further shows that HALMES may introduce new degradation for samples that originally benefited from CD. Overall, HALMES significantly reduces the risk caused by indiscriminate intervention, while the selective triggering mechanism could be further refined to better avoid false activations and unnecessary interventions.


\textbf{Relationship Between Intervention Frequency and Hallucination Mitigation.  }
To investigate whether HALMES's improvements arise from targeted intervention rather than merely reducing CD activation frequency, we conducted controlled experiments on 100 randomly sampled COCO images disjoint from the training set. Specifically, we considered two alternative activation strategies: random activation, which randomly determines whether to apply CD at each decoding step, and periodic activation, which applies CD over consecutive steps within a repeating 10-step cycle. We first analyzed the intervention statistics of HALMES and found that CD correction is activated at approximately 70\% of decoding steps. Based on this observation, we evaluated both random and periodic activation strategies at ratios of 30\%, 50\%, and 70\%, with the 70\% settings serving as frequency-comparable baselines. All results are averaged across the five evaluated CD methods.

\begin{wraptable}{r}{0.65\linewidth}
\vspace{-0.6\baselineskip}
\centering
\caption{Comparison with random and periodic CD activation on a COCO subset. Results are averaged across the five CD methods. \textbf{Bold} denotes the best result.}
\label{tab:activation_comparison}
\vspace{2pt}

{\scriptsize
\setlength{\tabcolsep}{2pt}
\renewcommand{\arraystretch}{1.05}

\begin{tabular}{@{}l|c|c|ccc|ccc|c@{}}
\toprule
\multirow{2}{*}{\textbf{Metric}}
& \multicolumn{1}{c|}{\cellcolor{groupcolorA}\textbf{Base}}
& \multicolumn{1}{c|}{\cellcolor{groupcolorA}\textbf{Full-stage}}
& \multicolumn{3}{c|}{\cellcolor{groupcolorA}\textbf{Random}}
& \multicolumn{3}{c|}{\cellcolor{groupcolorA}\textbf{Periodic}}
& \multicolumn{1}{c}{\cellcolor{groupcolorB}\textbf{Ours}} \\

\cmidrule(lr){2-2}
\cmidrule(lr){3-3}
\cmidrule(lr){4-6}
\cmidrule(lr){7-9}
\cmidrule(lr){10-10}

& \cellcolor{groupcolorA!40}Nucleus
& \cellcolor{groupcolorA!40}CD
& \cellcolor{groupcolorA!40}30\%
& \cellcolor{groupcolorA!40}50\%
& \cellcolor{groupcolorA!40}70\%
& \cellcolor{groupcolorA!40}3:7
& \cellcolor{groupcolorA!40}5:5
& \cellcolor{groupcolorA!40}7:3
& \cellcolor{groupcolorB!20}\textbf{HALMES} \\

\midrule

C$_I$$\downarrow$
& 14.45 & 13.11
& 13.90 & 13.90 & 14.60
& 14.18 & 14.28 & 14.67
& \cellcolor{groupcolorB!20}\textbf{12.29} \\

C$_S$$\downarrow$
& 37.02 & 36.44
& 36.90 & 36.63 & 36.85
& 35.63 & 36.86 & 36.48
& \cellcolor{groupcolorB!20}\textbf{32.40} \\

Cov.$\uparrow$
& 38.77 & 40.68
& 40.86 & 42.71 & 40.80
& 41.06 & 41.60 & 41.03
& \cellcolor{groupcolorB!20}\textbf{44.81} \\

\bottomrule
\end{tabular}
}

\vspace{-0.5\baselineskip}
\end{wraptable}

As shown in Table~\ref{tab:activation_comparison}, neither random nor periodic activation at 70\% yields notable improvements over full-stage CD, despite reducing the intervention frequency by approximately 30\%. In particular, both strategies increase C$_I$ from 13.11 to 14.60 and 14.67, respectively, while showing limited differences in C$_S$ and Coverage. In contrast, HALMES achieves the best performance across all three metrics, substantially outperforming both frequency-comparable baselines. Specifically, HALMES reduces C$_I$/C$_S$ by 2.31/4.45 and 2.38/4.08 percentage points compared with the frequency-comparable random and periodic activation, respectively, while improving Coverage by 4.01 and 3.78 points. These results indicate that simply reducing CD activation frequency does not consistently improve hallucination mitigation. Instead, they further support the association between hallucination risk and internal attention patterns, highlighting the importance of identifying anomalous token segments and applying targeted intervention during generation.

\section{Conclusion}
In this work, we propose HALMES, a lightweight plug-and-play module that detects hallucination-prone token segments from internal attention features during LVLM inference and selectively triggers contrastive decoding, converting indiscriminate full-stage intervention into on-demand correction. Experiments across different LVLM architectures and benchmarks show that HALMES improves existing contrastive decoding methods in most settings while reducing their negative impact on normal outputs. We also observe performance variations across datasets and decoding methods, indicating that reducing false triggers and improving cross-scenario robustness remain important future directions. Overall, these findings highlight the value of shifting inference-time hallucination mitigation from full-stage intervention to fine-grained anomalous-segment-level intervention.

\section*{Acknowledgments and Disclosure of Funding}
This work was supported in part by the New Generation Artificial Intelligence-National Science and Technology Major Project under Grant 2025ZD0123701, in part by the National Natural Science Foundation of China under Grants 62672348, 62476202, and 62272343, in part by the Explorers Program of Shanghai (Basic Research Funding) from Shanghai Municipal Science and Technology Commission under Grant 26TS1403800, in part by the Innovation Ecosystem Development Program of Shanghai Municipal Science and Technology Commission under Grant 26510761200, and in part by the Fundamental Research Funds for the Central Universities.

{
    \small
    \bibliographystyle{plainnat}
    \bibliography{main}
}

\newpage
\section*{Appendix}
In this appendix, we provide more details about the implementation, training recipe, and inference strategy of HALMES to complement the main paper. Additional benchmark descriptions, evaluation metrics, experimental results, and discussion are also incorporated. Below is the table of contents.

\noindent\textbf{A. Model}
\begin{enumerate}[leftmargin=3.5em, label=\arabic*., itemsep=0.15em, topsep=0.3em]
    \item Implementation Details
    \item Training Recipe
    \item Inference Strategy
\end{enumerate}

\noindent\textbf{B. Experiments}
\begin{enumerate}[leftmargin=3.5em, label=\arabic*., itemsep=0.15em, topsep=0.3em]
    \item Baselines
    \item Benchmarks \& Evaluation Metrics
    \item More Experimental Results
    \item Qualitative Results
\end{enumerate}

\noindent\textbf{C. Discussion}
\begin{enumerate}[leftmargin=3.5em, label=\arabic*., itemsep=0.15em, topsep=0.3em]
    \item Limitations \& Future Work
    \item Potential Impacts
\end{enumerate}

\renewcommand{\thesection}{\Alph{section}}
\renewcommand{\thesubsection}{\thesection.\arabic{subsection}}
\setcounter{section}{0}

\section{Model}
\subsection{Implementation Details}
The hallucination detector in HALMES is implemented as a 2-layer MLP to predict the hallucination risk of token segments. Specifically, the detector first maps the input segment representation into a 256-dimensional hidden space through a linear layer, followed by LayerNorm, GELU activation, and Dropout for normalization, nonlinear transformation, and regularization. A residual MLP block is further introduced to enhance nonlinear modeling capacity. Finally, a linear classification head outputs a scalar logit, which is converted into a hallucination risk score through a sigmoid function.
For a token segment of length $L$, its input attention features are denoted as $\mathbf{X}\in\mathbb{R}^{L\times D}$, where $D=1156$ is the feature dimension of each token. Specifically, VAR is represented as a $36\times32$ layer-head matrix and flattened into a 1152-dimensional vector, while the remaining four attention-based features, including prompt attention, generated-text attention, text-visual ratio, and outgoing attention entropy, are scalar values. Since different samples contain token segments of varying lengths, we aggregate $\mathbf{X}$ along the token dimension using five statistics: full-segment mean, tail mean over the last 5 tokens, standard deviation, temporal slope, and maximum value. These statistics are concatenated into a fixed-dimensional segment representation with dimension $5D=5780$, and the MLP outputs a single value indicating the probability that the token segment is hallucination-prone.


\subsection{Training Recipe}
To construct the training data for the hallucination abnormal segment detector, we perform offline data generation on a single NVIDIA A800-SXM4-80GB GPU. Specifically, we run the base LVLM on 5,000 randomly sampled images from the MSCOCO train2014 dataset \cite{lin2014microsoft} and record attention-related internal features at each token decoding step during caption generation. We then extract nouns from the generated captions and determine whether each noun refers to a non-existent object in the image according to the CHAIR protocol. For each noun, we locate its token id in the generated sequence and construct a token bag using the historical token features preceding that noun. Thus, each response yields as many token bags as its extracted nouns. The bag label is determined by whether the corresponding noun is hallucinated. In this way, the training data explicitly models the generation context before hallucination occurs, enabling the detector to learn abnormal token-segment patterns preceding hallucinated objects. During training, we split the dataset into training and validation sets in a stratified manner according to positive and negative labels, and adopt class reweighting to alleviate the imbalance between hallucinated and non-hallucinated samples. The model is optimized with AdamW, using a linear warmup followed by cosine learning-rate decay. We further employ mixed precision and gradient clipping to improve stability. During validation, AUROC is used as the primary criterion for model selection, and the checkpoint with the best validation performance is saved for real-time hallucination abnormal segment detection during inference.


\subsection{Inference Strategy}

\begin{algorithm}[t]
\caption{Inference-time deployment of HALMES}
\label{alg:halmes_inference}
\begin{algorithmic}[1]
\Require Feature stream $\{\mathbf z_t\}_{t=1}^{T}$, HALMES model $g_\phi$, window size $K$, prior statistics, thresholds $\theta_{\mathrm{high}}, \theta_{\mathrm{low}}$
\Ensure Intervention decisions $\{q_t\}_{t=1}^{T}$

\State Initialize feature buffer $\mathcal B \leftarrow \emptyset$, intervention state $q_0 \leftarrow 0$
\State Initialize warm-up counter $u \leftarrow 0$ and cool-down counter $c \leftarrow 0$

\For{$t=1$ to $T$}
    \State Append $\mathbf z_t$ to $\mathcal B$
    \If{statistics are uninitialized}
        \State Initialize statistics from priors and update once with $\mathbf z_t$
    \EndIf
    \State Construct $\mathbf Z_t$ from the last $\min(K,|\mathcal B|)$ features in $\mathcal B$
    \State Normalize $\mathbf Z_t$ to obtain $\hat{\mathbf Z}_t$, set $\mathbf{s}_t \leftarrow \mathrm{Agg}(\hat{\mathbf Z}_t)$, and compute $p_t \leftarrow \mathrm{sigmoid}(g_\phi(\mathbf{s}_t))$

    \If{$q_{t-1}=0$ and $p_t > \theta_{\mathrm{high}}$}
        \State Increase warm-up counter $u$ and reset cool-down counter $c \leftarrow 0$
        \State Set $q_t=1$ once $u$ reaches a preset limit; otherwise keep $q_t=0$
    \ElsIf{$q_{t-1}=1$ and $p_t < \theta_{\mathrm{low}}$}
        \State Increase cool-down counter $c$ and reset warm-up counter $u \leftarrow 0$
        \State Set $q_t=0$ once $c$ reaches a preset limit; otherwise keep $q_t=1$
    \Else
        \State Reset the inactive counter and keep the current intervention state
    \EndIf

    \If{$q_t=1$}
        \State Use the contrastive decoding branch
    \Else
        \State Use the original decoding branch
    \EndIf
\EndFor
\end{algorithmic}
\end{algorithm}

\textbf{Inference-time Deployment.} During inference, HALMES is integrated into the original LVLM decoding process as an external lightweight module. At each decoding step, we extract the attention-based feature vector from the current attention maps and append it to a feature buffer. To mitigate feature distribution shifts across different responses, we initialize the normalization statistics with the mean and variance computed from the training set, and perform a one-step Welford update using the first feature vector of the current response. The updated statistics are then fixed and used to normalize subsequent window features within the same response. Specifically, at each step, HALMES constructs a local token segment from the most recent $K$ feature vectors in the buffer and outputs the hallucination risk score of the current window. To avoid frequent activation or deactivation caused by local noise, we adopt a trend-aware double-threshold hysteresis mechanism. Based on the resulting intervention state, HALMES uses the corresponding contrastive decoding branch for high-risk segments and keeps the original decoding branch for normal segments. The complete inference procedure is shown in Algorithm~\ref{alg:halmes_inference}.


\textbf{Threshold Calibration.} The main inference hyperparameters include the local window size $K$, the activation threshold $\theta_{\mathrm{high}}$, and the deactivation threshold $\theta_{\mathrm{low}}$. Among them, $K$ is mainly determined according to the length distribution of token segments in the training data. Since HALMES learns abnormal patterns from token segments preceding hallucinated nouns during training, the inference window should remain close to the effective length of training samples. Meanwhile, to preserve early-warning capability, we set the local window size slightly smaller than the average effective length of training samples. After fixing $K$, we further calibrate $\theta_{\mathrm{high}}$ and $\theta_{\mathrm{low}}$ on the validation set according to the trade-off between Recall and false positive rate (FPR). Since HALMES aims to identify high-risk token segments before hallucinated objects are generated, high Recall is prioritized to reduce missed detections. At the same time, an excessively high FPR may falsely trigger intervention during normal generation and reintroduce the over-intervention problem. Therefore, we select a threshold combination that controls FPR while maintaining high Recall, and set $\theta_{\mathrm{low}}<\theta_{\mathrm{high}}$ to form a hysteresis interval, reducing frequent state switching caused by short-term score fluctuations. The effects of different threshold settings are reported in the Table~\ref{tab:threshold_selection}.


\section{Experiments}
\subsection{Baselines}
We select VCD \cite{leng2024mitigating}, M3ID \cite{favero2024multi}, AVISC \cite{woo2025dontmissforesttrees}, SID \cite{huo2024self}, and Self-Aug \cite{im2026selfaugqueryentropyadaptive} as the main baselines because the central goal of this work is to examine whether contrastive decoding should be applied selectively to hallucination-prone segments, rather than continuously throughout the entire generation process. HALMES is therefore positioned as a plug-and-play triggering module for existing CD methods, instead of a standalone adaptive decoding rule. For this reason, we focus on controlled comparisons within each CD backbone: we keep its original contrastive formulation, logit adjustment, and decoding rule unchanged, and compare the full-stage version with its HALMES-guided counterpart. This design isolates the contribution of the triggering mechanism and evaluates whether selective intervention can preserve the hallucination mitigation ability of contrastive decoding while reducing its side effects on normal generation trajectories.


\subsection{Benchmarks \& Evaluation Metrics}
\paragraph{Benchmarks.  }
We evaluate HALMES on four datasets, namely MSCOCO \cite{lin2014microsoft}, AMBER \cite{wang2023amber}, MMHal-Bench \cite{sun2023aligninglargemultimodalmodels}, and Bingo \cite{cui2023holistic}. Following prior work \cite{chen2024halc, duan2025truthprint, huang2024opera}, we randomly sample 500 examples from MSCOCO val2014 for evaluation. For AMBER, we use the first 300 samples in our experiments. MSCOCO and AMBER mainly assess object hallucination in open-ended generation based on object existence and visual factual consistency. MMHal-Bench covers tasks such as fine-grained visual understanding, relational reasoning, scene-level judgment, and open-ended description, evaluating both the informativeness and visual faithfulness of model responses. Bingo further examines model robustness from the perspectives of regional bias, OCR bias, factual bias, and image/text-induced interference. Together, these datasets allow us to evaluate both the hallucination mitigation capability of HALMES and its robustness under complex evaluation scenarios.


\paragraph{Evaluation Metrics.}
We adopt the official evaluation protocol and metrics for each benchmark and report them separately. Specifically, MSCOCO and AMBER follow their standard object-level hallucination protocols, which evaluate object mentions in generated responses against image annotations, while MMHal-Bench and Bingo follow their original LLM-based judging protocols.

For MSCOCO, we report CHAIR$_i$, CHAIR$_s$, and Cov. Following the original CHAIR definition, CHAIR$_i$ measures the fraction of hallucinated object mentions among all mentioned objects, while CHAIR$_s$ measures the fraction of generated sentences containing at least one hallucinated object:
\begin{equation}
    \mathrm{CHAIR}_i =
    \frac{
    |\{\mathrm{hallucinated\ objects}\}|
    }{
    |\{\mathrm{all\ objects\ mentioned}\}|
    },
\end{equation}
\begin{equation}
    \mathrm{CHAIR}_s =
    \frac{
    |\{\mathrm{sentences\ with\ hallucinated\ object}\}|
    }{
    |\{\mathrm{all\ sentences}\}|
    }.
\end{equation}
The hallucinated objects are object mentions in generated captions that are not supported by the corresponding images. Lower CHAIR$_i$ and CHAIR$_s$ indicate fewer object hallucinations.
We additionally report coverage (Cov.) to measure the coverage of ground-truth objects:
\begin{equation}
    \mathrm{Cov} =
    \frac{
    |\{\mathrm{unique\ ground\text{-}truth\ objects\ correctly\ mentioned}\}|
    }{
    |\{\mathrm{all\ ground\text{-}truth\ objects}\}|
    },
\end{equation}
where $\{\mathrm{unique\ ground\text{-}truth\ objects\ correctly\ mentioned}\}$ denotes the distinct ground-truth objects correctly mentioned in generated responses, and $\{\mathrm{all\ ground\text{-}truth\ objects}\}$ denotes all ground-truth objects in the evaluated images. A higher Cov. indicates better object coverage.

For AMBER, we report CHAIR, Cover, Hal, and Cog on the generative task. Following the original AMBER protocol, given a model response $R$, nouns are first extracted to obtain the initial object set $R_{\mathrm{obj}}$, which is then filtered by the full annotated object list $X_{\mathrm{obj}}$:
\begin{equation}
    R'_{\mathrm{obj}} = R_{\mathrm{obj}} \cap X_{\mathrm{obj}} .
\end{equation}
Let $A_{\mathrm{obj}}$ denote the set of annotated objects that truly appear in the image, $H_{\mathrm{obj}}$ denote the predefined hallucinatory target object set, and $\mathrm{len}(\cdot)$ denote the number of elements in a set. The four generative metrics in AMBER are defined as:
\begin{equation}
    \mathrm{CHAIR}(R)
    =
    1 -
    \frac{
    \mathrm{len}(R'_{\mathrm{obj}} \cap A_{\mathrm{obj}})
    }{
    \mathrm{len}(R'_{\mathrm{obj}})
    },
\end{equation}
\begin{equation}
    \mathrm{Cover}(R)
    =
    \frac{
    \mathrm{len}(R'_{\mathrm{obj}} \cap A_{\mathrm{obj}})
    }{
    \mathrm{len}(A_{\mathrm{obj}})
    },
\end{equation}
\begin{equation}
    \mathrm{Hal}(R)
    =
    \begin{cases}
    1, & \text{if } \mathrm{CHAIR}(R) \neq 0, \\
    0, & \text{otherwise},
    \end{cases}
\end{equation}
\begin{equation}
    \mathrm{Cog}(R)
    =
    \frac{
    \mathrm{len}(R'_{\mathrm{obj}} \cap H_{\mathrm{obj}})
    }{
    \mathrm{len}(R'_{\mathrm{obj}})
    }.
\end{equation}
CHAIR and Hal quantify hallucination severity at the object and response levels, respectively. Cover measures ground-truth object coverage and Cog reflects the tendency to generate predefined hallucination-prone objects. Better performance is reflected by lower CHAIR, Hal, and Cog values and a higher Cover value. Consistent with the original AMBER protocol, each metric is obtained by averaging the corresponding response-level scores over the test set.

For MMHal-Bench and Bingo, we adopt their original LLM-based evaluation protocols, with GPT-4o used as the judge model in our implementation. MMHal-Bench reports Hal and Score, where Hal measures the proportion of responses judged to contain hallucinations, and Score reflects the average response quality in terms of informativeness and visual faithfulness. Bingo reports Score to evaluate response reliability under regional bias, OCR bias, factual bias, image-to-image interference, and text-to-image interference. The reported scores for both benchmarks are averaged over test samples.

\subsection{More Experimental Results}
\subsubsection{How can the intervention threshold be selected?}
\begin{figure}[H]
    \centering
    \begin{subfigure}[t]{0.48\textwidth}
        \centering
        \includegraphics[width=\linewidth]{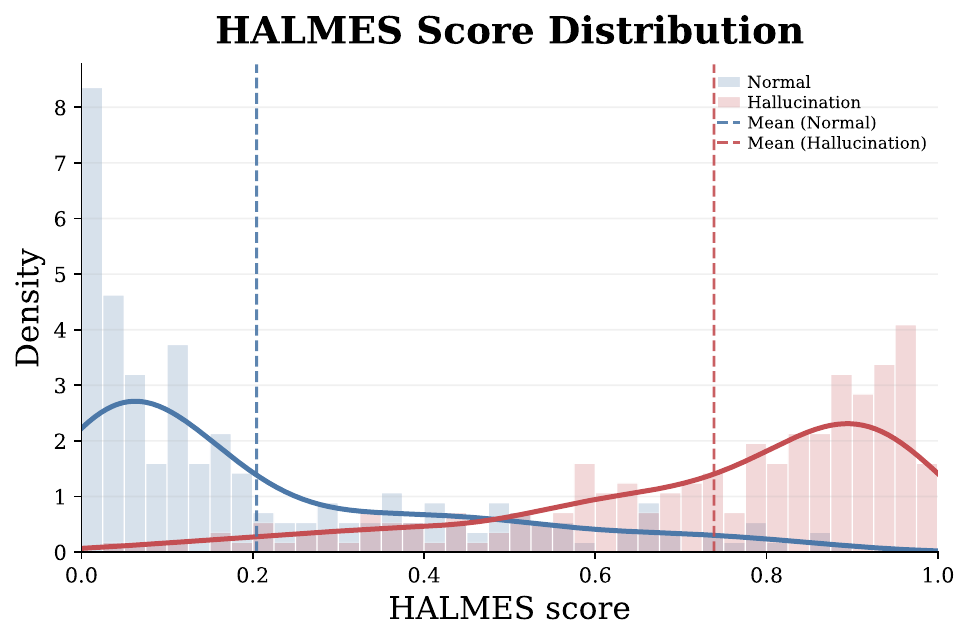}
        \caption{Detector score distribution.}
        \label{fig:threshold_recall_fpr}
    \end{subfigure}
    \hfill
    \begin{subfigure}[t]{0.48\textwidth}
        \centering
        \includegraphics[width=\linewidth]{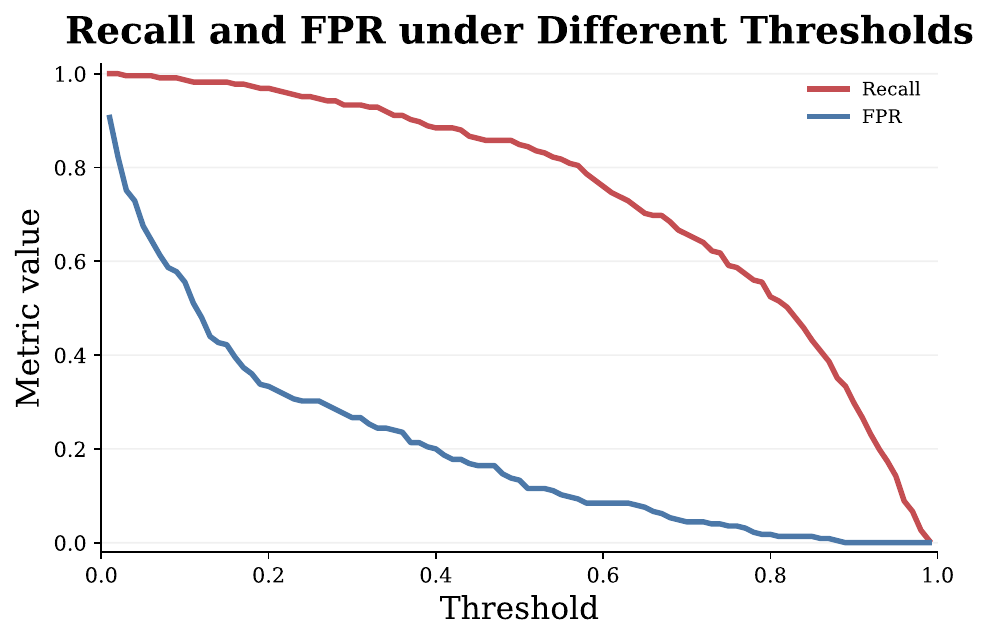}
        \caption{Recall--FPR trade-off.}
        \label{fig:threshold_score_dist}
    \end{subfigure}
    \caption{
    Threshold selection analysis for HALMES. The score distribution shows the separability between normal and hallucination-prone segments, while the Recall--FPR curve illustrates the trade-off between detecting hallucination-prone segments and avoiding false activations.
    }
    \label{fig:threshold_analysis}
\end{figure}

To analyze the effect of threshold selection on detection performance, we consider multiple threshold configurations from two aspects: the FPR budget $\alpha$ and the hysteresis gap $\theta_{\mathrm{high}}-\theta_{\mathrm{low}}$. For each $\alpha$, the threshold pair is selected on the validation set according to the following criterion:
\begin{equation}
    (\theta_{\mathrm{high}}^{*}, \theta_{\mathrm{low}}^{*})
    =
    \arg\max_{\theta_{\mathrm{high}},\,\theta_{\mathrm{low}}}
    \mathrm{Recall},
    \quad
    \mathrm{s.t.}\quad
    \mathrm{FPR} \leq \alpha,
\end{equation}
where $\theta_{\mathrm{high}}$ and $\theta_{\mathrm{low}}$ denote the high and low thresholds for activating and deactivating intervention, respectively. As illustrated in Fig.~\ref{fig:threshold_analysis}, increasing sensitivity improves Recall but also raises the risk of false activations on normal segments. Therefore, the FPR constraint is used to limit false activations and reduce the side effects caused by unnecessary intervention. Among the threshold pairs satisfying this constraint, we choose the one with the highest Recall to cover as many hallucination-prone segments as possible. This strategy therefore selects the most sensitive detector under a controlled FPR budget.

To cover different threshold behaviors, we consider the following configurations:

\begin{itemize}[leftmargin=1.5em, itemsep=0.15em, topsep=0.2em]
    \item \textbf{FPR budget.} We first fix the hysteresis gap at $0.10$ and vary $\alpha \in \{0.05, 0.10, 0.15\}$ to examine the Recall--FPR trade-off under different levels of FPR control. These three settings correspond to conservative, balanced, and relaxed triggering constraints, respectively.

   \item \textbf{Hysteresis gap.} Based on the selected FPR budget from the above comparison, we further vary the hysteresis gap $\theta_{\mathrm{high}}-\theta_{\mathrm{low}} \in \{0, 0.05, 0.15\}$. The zero-gap setting corresponds to single-threshold triggering without hysteresis, while larger gaps introduce wider activation--deactivation intervals to improve triggering stability under local score fluctuations.
\end{itemize}

\begin{table}[H]
\centering
\caption{Threshold selection analysis on a held-out AMBER calibration subset using Qwen3VL+VCD, disjoint from the final test samples. The first block varies the FPR budget $\alpha$ with the hysteresis gap fixed at $0.10$, while the second block varies the hysteresis gap with the selected FPR budget fixed at $\alpha=0.05$. \textbf{Bold} denotes the configuration used in our main experiments.}
\label{tab:threshold_selection}
\vspace{2pt}
\resizebox{0.62\textwidth}{!}{%
\begin{tabular}{l|cc|cccc}
\toprule
\textbf{Setting} & $\alpha$ & Gap
& CHAIR$\downarrow$ & Cover$\uparrow$ & Cog$\downarrow$ & Hal$\downarrow$ \\
\midrule

\rowcolor{groupcolorA}
\multicolumn{7}{l}{\textit{\textbf{Varying FPR budget}}} \\
\textbf{Conservative budget} & \textbf{0.05} & \textbf{0.10} 
& \textbf{7.25} & 49.36 & 3.78 & \textbf{31.00} \\
Balanced budget     & 0.10 & 0.10 & 9.51 & 49.48 & 4.52 & 46.00 \\
Relaxed budget      & 0.15 & 0.10 & 10.68 & 50.83 & 4.35 & 49.00 \\

\midrule
\rowcolor{groupcolorA}
\multicolumn{7}{l}{\textit{\textbf{Varying hysteresis gap}}} \\
No hysteresis        & 0.05 & 0.00 & 8.53 & 48.77 & \textbf{3.42} & 41.00 \\
Narrow gap           & 0.05 & 0.05 & 8.81 & \textbf{51.81} & 4.14 & 45.00 \\
Wide gap             & 0.05 & 0.15 & 8.92 & 50.68 & 3.67 & 46.00 \\

\bottomrule
\end{tabular}%
}
\end{table}

As shown in Table~\ref{tab:threshold_selection}, the conservative budget provides the most reliable hallucination reduction, achieving the lowest CHAIR and Hal among all settings. In comparison, larger FPR budgets introduce more aggressive triggering and increase hallucination-related errors, while alternative hysteresis gaps do not further improve the hallucination metrics. Therefore, to preserve the hallucination mitigation effect and maintain a low false-triggering tendency, we use the threshold pair selected under the conservative budget as the default setting.


\subsubsection{Can a hallucination detector trained on one LVLM architecture be transferred to another LVLM architecture?}
To examine cross-architecture transferability, we compare in-domain detectors with cross-model evaluation. As shown in Table~\ref{tab:cross_model_detector}, although the transferred detectors still retain moderate AUROC values, their performance is clearly lower than in-domain evaluation. Specifically, AUROC drops from above $90$ to $80.02/85.67$, while FPR increases substantially to $34.22/33.63$, indicating that cross-model detectors introduce more false activations. Interestingly, the Qwen-to-InternVL setting achieves a high Recall of $92.86$, but this is accompanied by a high FPR of $33.63$. This suggests that the transferred detector tends to over-trigger on the target model rather than accurately separating hallucination-prone and normal segments. These results indicate that different LVLMs exhibit distinct attention distributions and decision boundaries, which limits the reliability of direct cross-model detector transfer.

\begin{table}[t]
\centering
\caption{Cross-architecture evaluation of HALMES detectors. Base denotes in-domain evaluation, where the detector is trained and tested on the same LVLM architecture. Cross-architecture rows evaluate whether a detector trained on one LVLM can be transferred to another.}
\label{tab:cross_model_detector}
\vspace{2pt}
\resizebox{0.78\textwidth}{!}{%
\begin{tabular}{ll|ccccc}
\toprule
\textbf{Train LVLM} & \textbf{Test LVLM}
& AUROC$\uparrow$ & Accuracy$\uparrow$ & Recall$\uparrow$ & PR-AUC$\uparrow$ & FPR$\downarrow$ \\
\midrule
\rowcolor{groupcolorA}
\multicolumn{7}{l}{\textit{\textbf{In-domain evaluation}}} \\
Qwen3VL-8B     & Qwen3VL-8B     & 91.31 & 86.73 & 86.39 & 88.76 & 12.93 \\
InternVL3.5-8B & InternVL3.5-8B & 90.52 & 83.63 & 87.20 & 90.03 & 19.94 \\
\midrule
\rowcolor{groupcolorA}
\multicolumn{7}{l}{\textit{\textbf{Cross-architecture evaluation}}} \\
InternVL3.5-8B & Qwen3VL-8B     & 80.02 & 74.22 & 82.67 & 77.46 & 34.22 \\
Qwen3VL-8B     & InternVL3.5-8B & 85.67 & 79.61 & 92.86 & 83.30 & 33.63 \\
\bottomrule
\end{tabular}%
}
\end{table}

\subsubsection{How does deploying the HALMES affect inference speed during contrastive decoding?}
\begin{figure}[H]
    \centering
    \includegraphics[width=0.88\textwidth]{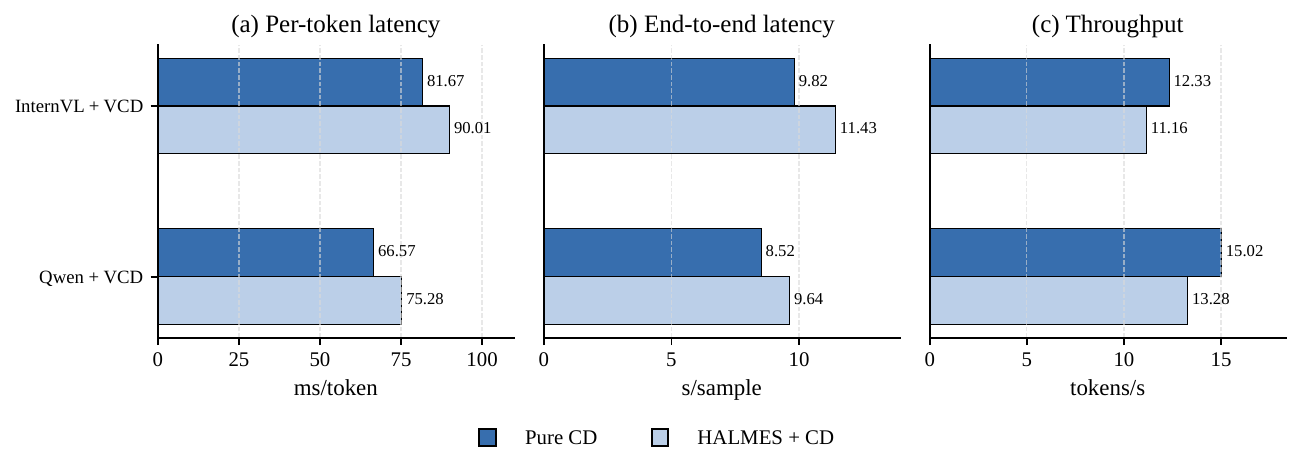}
    \caption{
    Inference speed comparison between Pure CD and HALMES + CD. 
    All results are measured on the same hardware over 100 test samples, with per-token latency, end-to-end latency, and throughput reported to quantify the overhead introduced by the HALMES detector.
    }
    \label{fig:speed}
\end{figure}

To evaluate the impact of the MLP detector on inference efficiency, we compare Pure CD and HALMES + CD in Fig.~\ref{fig:speed}. We report per-token latency, end-to-end latency, and throughput, corresponding to the decoding time per generated token, the total time per sample, and the number of generated tokens per second, respectively. HALMES + CD introduces only a slight slowdown across these metrics, since the detector is a lightweight MLP with marginal forward cost compared with LVLM decoding. These results indicate that HALMES enables selective hallucination mitigation with acceptable inference overhead.

\subsubsection{How do the attention-based features contribute to hallucination detection and mitigation?}

\textbf{Hallucination Detection.}
To evaluate the contribution of individual attention-based features, we conduct leave-one-feature-out ablations on both LVLMs while keeping the detector architecture and training settings unchanged. As shown in Table~\ref{tab:feature_ablation_detection}, removing any feature decreases AUROC and increases FPR on both models, indicating that all five features contribute to hallucination-risk detection. Among them, VAR plays the most important role, as its removal decreases AUROC by 12.37 and 8.80 percentage points on Qwen3VL-8B and InternVL3.5-8B, respectively, while substantially increasing FPR. Although certain ablations slightly improve individual metrics such as Recall or PR-AUC, the full feature set achieves the highest AUROC and lowest FPR on both architectures. These results suggest that the proposed attention-based features provide complementary information for identifying hallucination-prone token segments.

\begin{table}[t]
\centering
\caption{Component-wise ablation of the attention-based features for hallucination detection. All metrics are reported as percentages. \textbf{Bold} denotes the best result for each LVLM.}
\label{tab:feature_ablation_detection}
\vspace{2pt}
\resizebox{0.86\textwidth}{!}{%
\begin{tabular}{l|ccccc}
\toprule
\textbf{Setting}
& AUROC$\uparrow$
& Accuracy$\uparrow$
& Recall$\uparrow$
& PR-AUC$\uparrow$
& FPR$\downarrow$ \\
\midrule
\rowcolor{groupcolorA}
\multicolumn{6}{l}{\textit{\textbf{Qwen3VL-8B}}} \\
w/o VAR
& 78.94 & 73.11 & 81.78 & 76.53 & 35.56 \\
w/o Prompt Attention
& 88.68 & 80.22 & 75.56 & 88.42 & 15.11 \\
w/o Generated-Text Attention
& 88.80 & 82.22 & 84.89 & \textbf{89.54} & 20.44 \\
w/o Text-Visual Ratio
& 87.87 & 82.22 & 82.67 & 88.81 & 18.22 \\
w/o Outgoing Attention Entropy
& 87.91 & 82.44 & 80.89 & 88.89 & 16.00 \\
Full
& \textbf{91.31} & \textbf{86.73} & \textbf{86.39} & 88.76 & \textbf{12.93} \\
\midrule
\rowcolor{groupcolorA}
\multicolumn{6}{l}{\textit{\textbf{InternVL3.5-8B}}} \\
w/o VAR
& 81.72 & 75.30 & \textbf{88.10} & 79.28 & 37.50 \\
w/o Prompt Attention
& 88.14 & 81.70 & 85.71 & 88.63 & 22.32 \\
w/o Generated-Text Attention
& 88.07 & 81.85 & 87.80 & 88.68 & 24.11 \\
w/o Text-Visual Ratio
& 88.15 & 81.70 & 77.98 & 88.71 & 21.13 \\
w/o Outgoing Attention Entropy
& 88.13 & 80.95 & 80.36 & 88.58 & 21.96 \\
Full
& \textbf{90.52} & \textbf{83.63} & 87.20 & \textbf{90.03} & \textbf{19.94} \\
\bottomrule
\end{tabular}%
}
\end{table}

\textbf{Hallucination Mitigation.}
We further integrate each feature-ablated detector with the five CD methods to examine how individual features affect hallucination mitigation. The averaged results are reported in Table~\ref{tab:feature_ablation_mitigation}. The full feature set consistently achieves the lowest C$_I$ and C$_S$ on both LVLMs, outperforming all feature-ablated variants. Removing any feature leads to increased hallucination metrics, although the extent of degradation varies across features and architectures. In particular, removing the Text-Visual Ratio substantially weakens mitigation performance on InternVL3.5-8B, while removing VAR also considerably reduces the improvements obtained by HALMES. Meanwhile, Coverage remains relatively stable on Qwen3VL-8B but exhibits a more noticeable trade-off on InternVL3.5-8B. Overall, these findings further demonstrate the complementary contributions of the five attention-based features, supporting their joint use for effective hallucination detection and targeted intervention.

\begin{table}[htbp]
\centering
\caption{Component-wise ablation of the attention-based features for hallucination mitigation on a randomly sampled 100-image COCO subset disjoint from the training set. Results are averaged across the five CD methods. All metrics are reported as percentages. \textbf{Bold} denotes the best result among HALMES variants for each LVLM.}
\label{tab:feature_ablation_mitigation}
\vspace{2pt}
\resizebox{0.86\textwidth}{!}{%
\begin{tabular}{l|ccc|ccc}
\toprule
\multirow{2}{*}{\textbf{Setting}}
& \multicolumn{3}{c|}{\textbf{Qwen3VL-8B}}
& \multicolumn{3}{c}{\textbf{InternVL3.5-8B}} \\
\cmidrule(lr){2-4}\cmidrule(lr){5-7}
& C$_I$$\downarrow$ & C$_S$$\downarrow$ & Cov.$\uparrow$
& C$_I$$\downarrow$ & C$_S$$\downarrow$ & Cov.$\uparrow$ \\
\midrule
\rowcolor{groupcolorA}
\multicolumn{7}{l}{\textit{\textbf{Base}}} \\
Nucleus
& 12.50 & 28.00 & 56.00
& 12.31 & 36.00 & 68.00 \\
\midrule
\rowcolor{groupcolorA}
\multicolumn{7}{l}{\textit{\textbf{Full-Stage Contrastive Decoding}}} \\
Full-stage CD
& 11.71 & 29.40 & 54.16
& 11.87 & 38.20 & 67.92 \\
\midrule
\rowcolor{groupcolorB}
\multicolumn{7}{l}{\textit{\textbf{HALMES (Feature Ablations)}}} \\
w/o VAR
& 11.30 & 26.80 & 55.04
& 12.13 & 37.60 & 66.96 \\
w/o Prompt Attention
& 11.38 & 27.20 & \textbf{55.76}
& 11.57 & 36.40 & 66.96 \\
w/o Generated-Text Attention
& 10.14 & 26.40 & \textbf{55.76}
& 10.82 & 34.80 & 66.88 \\
w/o Text-Visual Ratio
& 11.37 & 28.00 & 55.36
& 12.64 & 38.40 & 67.76 \\
w/o Outgoing Attention Entropy
& 10.27 & 24.60 & 54.24
& 11.63 & 36.40 & \textbf{68.48} \\
Full
& \textbf{9.11} & \textbf{24.40} & 55.68
& \textbf{9.98} & \textbf{30.20} & 66.48 \\
\bottomrule
\end{tabular}%
}
\end{table}

\subsubsection{Can HALMES preserve general LVLM capabilities beyond object hallucinations?}

To further investigate the impact of HALMES on general LVLM capabilities, we conduct additional evaluations on MME \cite{fu2023mme}, covering object existence, counting, position, color, OCR, and reasoning. To better accommodate token-segment-level detection, we adopt complete-response prompting to encourage sufficiently detailed responses during generation. As shown in Table~\ref{tab:mme_general}, HALMES improves the overall MME score across all ten combinations of LVLMs and CD methods, with maximum relative gains of 4.14\% on Qwen3VL-8B and 2.63\% on InternVL3.5-8B. Notably, improvements are observed in reasoning for nine of the ten settings, together with considerable gains in position understanding. Although some categories exhibit slight degradation, these results suggest that HALMES can largely preserve and potentially enhance broader LVLM capabilities while mitigating object hallucinations, further indicating the potential extensibility of the proposed targeted intervention paradigm.

\begin{table}[t]
\centering
\caption{Relative MME score changes (\%) of HALMES over the corresponding full-stage CD methods under complete-response prompting. Positive values indicate improvements.}
\label{tab:mme_general}
\vspace{2pt}
\resizebox{0.86\textwidth}{!}{%
\begin{tabular}{l|cccccc|c}
\toprule
\textbf{CD Method}
& Exist.
& Count
& Position
& Color
& OCR
& Reasoning
& Total \\
\midrule
\rowcolor{groupcolorA}
\multicolumn{8}{l}{\textit{\textbf{Qwen3VL-8B}}} \\
VCD
& 0.00 & +2.86 & +6.74 & -5.00 & +4.41 & +3.07 & +1.67 \\
M3ID
& 0.00 & 0.00 & +9.78 & +2.70 & +8.45 & +5.31 & +4.14 \\
AVISC
& 0.00 & 0.00 & +9.78 & -2.56 & 0.00 & +8.56 & +2.22 \\
SID
& 0.00 & +3.74 & 0.00 & 0.00 & +4.23 & +4.45 & +2.00 \\
Self-Aug
& +2.63 & 0.00 & +13.95 & 0.00 & -4.23 & -1.28 & +1.45 \\
\midrule
\rowcolor{groupcolorA}
\multicolumn{8}{l}{\textit{\textbf{InternVL3.5-8B}}} \\
VCD
& +2.63 & +0.99 & +12.22 & -7.69 & +4.62 & +4.81 & +2.43 \\
M3ID
& +2.70 & -2.80 & +9.57 & -3.42 & 0.00 & +7.54 & +1.85 \\
AVISC
& 0.00 & +9.78 & -1.02 & -0.90 & +4.62 & +4.92 & +2.63 \\
SID
& 0.00 & +1.98 & -1.00 & +2.86 & -4.62 & +6.00 & +0.77 \\
Self-Aug
& +2.63 & +1.94 & +1.03 & +0.94 & -4.41 & +2.92 & +0.83 \\
\bottomrule
\end{tabular}%
}
\end{table}

\begin{table}[t]
\centering
\caption{Hallucination detection performance of HALMES on Qwen3.5-9B. All metrics are reported as percentages.}
\label{tab:qwen35_detection}
\vspace{2pt}
\resizebox{0.72\textwidth}{!}{%
\begin{tabular}{l|ccccc}
\toprule
\textbf{LVLM}
& AUROC$\uparrow$
& Accuracy$\uparrow$
& Recall$\uparrow$
& PR-AUC$\uparrow$
& FPR$\downarrow$ \\
\midrule
Qwen3.5-9B
& 88.18 & 81.15 & 82.38 & 88.44 & 19.09 \\
\bottomrule
\end{tabular}%
}
\end{table}

\begin{table}[htbp]
\centering
\caption{Object hallucination mitigation results of HALMES on Qwen3.5-9B over a COCO subset. All metrics are reported as percentages. \textbf{Bold} denotes the best result.}
\label{tab:qwen35_mitigation}
\vspace{2pt}
\resizebox{0.5\textwidth}{!}{%
\begin{tabular}{l|ccc}
\toprule
\textbf{Method}
& C$_I$$\downarrow$
& C$_S$$\downarrow$
& Cov.$\uparrow$ \\
\midrule
\rowcolor{groupcolorA}
\multicolumn{4}{l}{\textit{\textbf{Base}}} \\
Nucleus
& 9.72 & 33.33 & 63.13 \\
\midrule
\rowcolor{groupcolorA}
\multicolumn{4}{l}{\textit{\textbf{Full-Stage Contrastive Decoding}}} \\
VCD
& 8.78 & 28.12 & 63.12 \\
M3ID
& 9.53 & 27.08 & 64.80 \\
AVISC
& 9.14 & 28.12 & 65.75 \\
SID
& 7.98 & 26.04 & 65.58 \\
Self-Aug
& 9.74 & 29.17 & 66.81 \\
\midrule
\rowcolor{groupcolorB}
\multicolumn{4}{l}{\textit{\textbf{HALMES (Ours)}}} \\
\rowcolor{groupcolorB!30}
HALMES+VCD
& 7.64 & 26.37 & 62.87 \\
\rowcolor{groupcolorB!30}
HALMES+M3ID
& 7.87 & 22.92 & 63.06 \\
\rowcolor{groupcolorB!30}
HALMES+AVISC
& \textbf{6.40} & \textbf{21.88} & \textbf{68.61} \\
\rowcolor{groupcolorB!30}
HALMES+SID
& 7.57 & 25.77 & 62.00 \\
\rowcolor{groupcolorB!30}
HALMES+Self-Aug
& 7.32 & 26.60 & 67.45 \\
\bottomrule
\end{tabular}%
}
\end{table}

\subsubsection{Can HALMES remain effective on a more native multimodal architecture?}

To further examine the applicability of HALMES to different LVLM architectures, we conduct additional experiments on Qwen3.5-9B, which adopts a hybrid attention architecture combining linear and full attention, distinct from the previously evaluated models. After lightweight model-specific detector training, HALMES achieves an AUROC of 88.18\% and an FPR of 19.09\%, as shown in Table~\ref{tab:qwen35_detection}, indicating that hallucination-related attention patterns remain detectable on this architecture. We further integrate HALMES with five CD methods and evaluate object hallucination mitigation on a COCO subset. As shown in Table~\ref{tab:qwen35_mitigation}, HALMES consistently reduces both CHAIR metrics across all five CD methods, although Coverage varies across settings. These results further demonstrate the effectiveness of HALMES on a more native multimodal architecture, supporting the broader applicability of attention-based hallucination detection and targeted intervention.

\subsection{Qualitative Results}
\begin{figure}[H]
    \centering
    \includegraphics[width=0.88\textwidth]{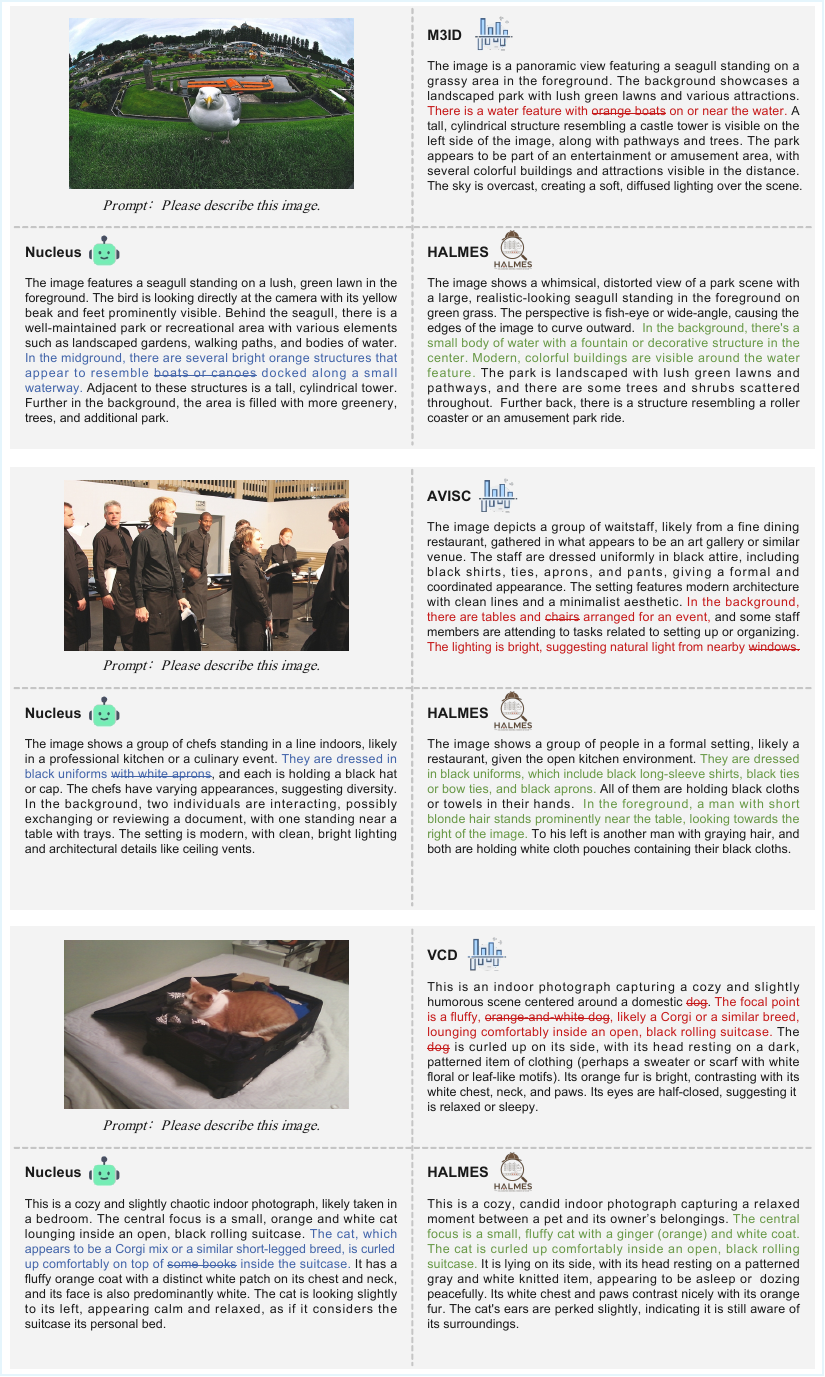}
    \caption{Qualitative comparison of M3ID, AVISC, VCD and their HALMES-guided variants.}
    \label{fig:qualitative_2}
\end{figure}

\begin{figure}[H]
    \centering
    \includegraphics[width=0.85\textwidth]{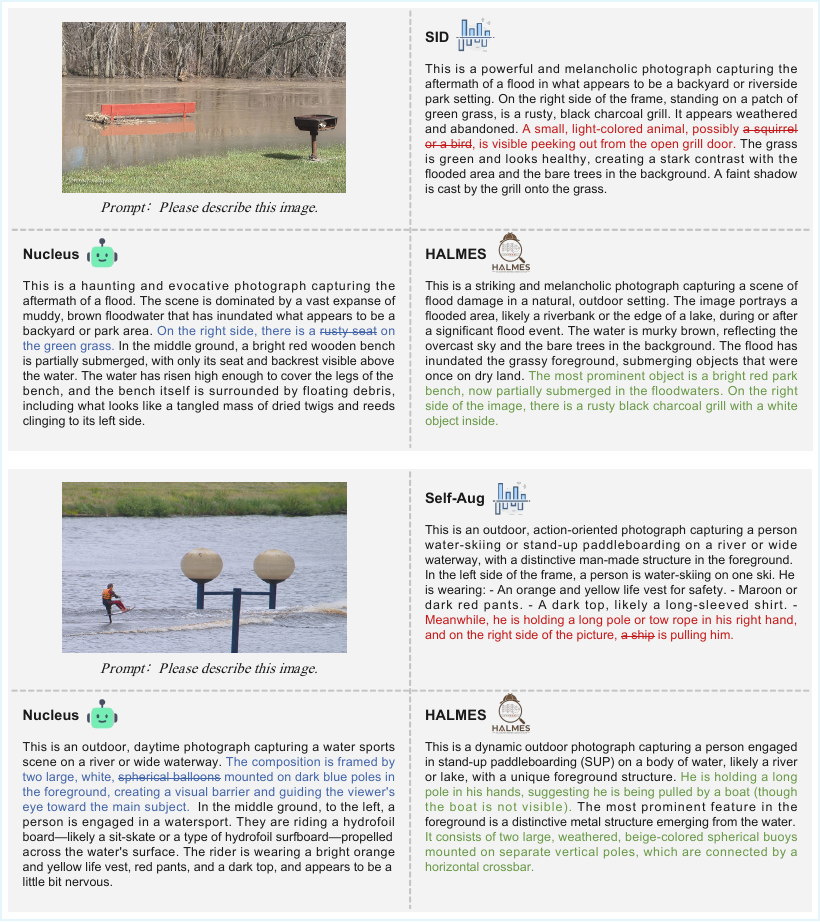}
    \caption{Qualitative comparison of SID, Self-Aug, and their HALMES-guided variants.}
    \label{fig:qualitative_1}
\end{figure}

Fig.~\ref{fig:qualitative_2} and Fig.~\ref{fig:qualitative_1} show qualitative comparisons between Pure CD and HALMES-guided CD. HALMES retains the hallucination mitigation effect of contrastive decoding while alleviating its side effects on normal generations, leading to more visually faithful and stable responses.

\section{Discussion}
\subsection{Limitations \& Future Work}
HALMES currently relies on predefined thresholds to determine whether contrastive decoding should be activated. Since different LVLM architectures, input scenarios, and generation stages may exhibit distinct attention distributions and risk-score scales, a fixed threshold may not capture the optimal decision boundary in all cases, potentially leading to false activations or missed interventions for certain samples. Therefore, future work should explore dynamic and adaptive thresholding mechanisms, enabling abnormality decisions to adjust according to the generation context and internal model states. In addition, we plan to investigate an end-to-end training framework that jointly optimizes hallucination-prone segment detection and contrastive decoding activation, thereby reducing the dependence on manually specified thresholds and further improving the robustness of HALMES across different models and complex scenarios.
\subsection{Potential Impacts}
This work introduces a selective hallucination mitigation framework for LVLMs, which converts full-stage implicit contrastive decoding into explicit on-demand intervention. It could potentially be used in multimodal assistant scenarios where faithful visual grounding and stable response generation are required, such as image captioning, visual question answering, and interactive vision-language applications. In scenarios where contrastive decoding is adopted for hallucination mitigation, HALMES can serve as a more targeted alternative by activating intervention only on hallucination-prone token segments. This may help reduce unnecessary perturbations to normal generation trajectories while retaining the mitigation ability of contrastive decoding. However, since HALMES relies on internal risk estimation rather than external factual verification, careful evaluation is still needed before applying it to safety-critical domains.


\clearpage

\clearpage

\newpage
\section*{NeurIPS Paper Checklist}

\begin{enumerate}

\item {\bf Claims}
    \item[] Question: Do the main claims made in the abstract and introduction accurately reflect the paper's contributions and scope?
    \item[] Answer: \answerYes{} 
    \item[] Justification: The abstract and introduction of this paper accurately reflect its contributions and scope.
    \item[] Guidelines:
    \begin{itemize}
        \item The answer \answerNA{} means that the abstract and introduction do not include the claims made in the paper.
        \item The abstract and/or introduction should clearly state the claims made, including the contributions made in the paper and important assumptions and limitations. A \answerNo{} or \answerNA{} answer to this question will not be perceived well by the reviewers. 
        \item The claims made should match theoretical and experimental results, and reflect how much the results can be expected to generalize to other settings. 
        \item It is fine to include aspirational goals as motivation as long as it is clear that these goals are not attained by the paper. 
    \end{itemize}

\item {\bf Limitations}
    \item[] Question: Does the paper discuss the limitations of the work performed by the authors?
    \item[] Answer: \answerYes{} 
    \item[] Justification: Yes, the paper discusses the limitations of the work in the appendix.
    \item[] Guidelines:
    \begin{itemize}
        \item The answer \answerNA{} means that the paper has no limitation while the answer \answerNo{} means that the paper has limitations, but those are not discussed in the paper. 
        \item The authors are encouraged to create a separate ``Limitations'' section in their paper.
        \item The paper should point out any strong assumptions and how robust the results are to violations of these assumptions (e.g., independence assumptions, noiseless settings, model well-specification, asymptotic approximations only holding locally). The authors should reflect on how these assumptions might be violated in practice and what the implications would be.
        \item The authors should reflect on the scope of the claims made, e.g., if the approach was only tested on a few datasets or with a few runs. In general, empirical results often depend on implicit assumptions, which should be articulated.
        \item The authors should reflect on the factors that influence the performance of the approach. For example, a facial recognition algorithm may perform poorly when image resolution is low or images are taken in low lighting. Or a speech-to-text system might not be used reliably to provide closed captions for online lectures because it fails to handle technical jargon.
        \item The authors should discuss the computational efficiency of the proposed algorithms and how they scale with dataset size.
        \item If applicable, the authors should discuss possible limitations of their approach to address problems of privacy and fairness.
        \item While the authors might fear that complete honesty about limitations might be used by reviewers as grounds for rejection, a worse outcome might be that reviewers discover limitations that aren't acknowledged in the paper. The authors should use their best judgment and recognize that individual actions in favor of transparency play an important role in developing norms that preserve the integrity of the community. Reviewers will be specifically instructed to not penalize honesty concerning limitations.
    \end{itemize}

\item {\bf Theory assumptions and proofs}
    \item[] Question: For each theoretical result, does the paper provide the full set of assumptions and a complete (and correct) proof?
    \item[] Answer: \answerYes{} 
    \item[] Justification: The paper provide the full set of assumptions and a complete (and correct) proof.
    \item[] Guidelines:
    \begin{itemize}
        \item The answer \answerNA{} means that the paper does not include theoretical results. 
        \item All the theorems, formulas, and proofs in the paper should be numbered and cross-referenced.
        \item All assumptions should be clearly stated or referenced in the statement of any theorems.
        \item The proofs can either appear in the main paper or the supplemental material, but if they appear in the supplemental material, the authors are encouraged to provide a short proof sketch to provide intuition. 
        \item Inversely, any informal proof provided in the core of the paper should be complemented by formal proofs provided in appendix or supplemental material.
        \item Theorems and Lemmas that the proof relies upon should be properly referenced. 
    \end{itemize}

    \item {\bf Experimental result reproducibility}
    \item[] Question: Does the paper fully disclose all the information needed to reproduce the main experimental results of the paper to the extent that it affects the main claims and/or conclusions of the paper (regardless of whether the code and data are provided or not)?
    \item[] Answer: \answerYes{} 
    \item[] Justification: The paper fully discloses all the necessary information to reproduce the main experimental results, ensuring that the main claims and conclusions can be independently verified.
    \item[] Guidelines:
    \begin{itemize}
        \item The answer \answerNA{} means that the paper does not include experiments.
        \item If the paper includes experiments, a \answerNo{} answer to this question will not be perceived well by the reviewers: Making the paper reproducible is important, regardless of whether the code and data are provided or not.
        \item If the contribution is a dataset and\slash or model, the authors should describe the steps taken to make their results reproducible or verifiable. 
        \item Depending on the contribution, reproducibility can be accomplished in various ways. For example, if the contribution is a novel architecture, describing the architecture fully might suffice, or if the contribution is a specific model and empirical evaluation, it may be necessary to either make it possible for others to replicate the model with the same dataset, or provide access to the model. In general. releasing code and data is often one good way to accomplish this, but reproducibility can also be provided via detailed instructions for how to replicate the results, access to a hosted model (e.g., in the case of a large language model), releasing of a model checkpoint, or other means that are appropriate to the research performed.
        \item While NeurIPS does not require releasing code, the conference does require all submissions to provide some reasonable avenue for reproducibility, which may depend on the nature of the contribution. For example
        \begin{enumerate}
            \item If the contribution is primarily a new algorithm, the paper should make it clear how to reproduce that algorithm.
            \item If the contribution is primarily a new model architecture, the paper should describe the architecture clearly and fully.
            \item If the contribution is a new model (e.g., a large language model), then there should either be a way to access this model for reproducing the results or a way to reproduce the model (e.g., with an open-source dataset or instructions for how to construct the dataset).
            \item We recognize that reproducibility may be tricky in some cases, in which case authors are welcome to describe the particular way they provide for reproducibility. In the case of closed-source models, it may be that access to the model is limited in some way (e.g., to registered users), but it should be possible for other researchers to have some path to reproducing or verifying the results.
        \end{enumerate}
    \end{itemize}

\item {\bf Open access to data and code}
    \item[] Question: Does the paper provide open access to the data and code, with sufficient instructions to faithfully reproduce the main experimental results, as described in supplemental material?
    \item[] Answer: \answerYes{} 
    \item[] Justification: The code and data have been submitted as supplementary material in ZIP format.
    \item[] Guidelines:
    \begin{itemize}
        \item The answer \answerNA{} means that paper does not include experiments requiring code.
        \item Please see the NeurIPS code and data submission guidelines (\url{https://neurips.cc/public/guides/CodeSubmissionPolicy}) for more details.
        \item While we encourage the release of code and data, we understand that this might not be possible, so \answerNo{} is an acceptable answer. Papers cannot be rejected simply for not including code, unless this is central to the contribution (e.g., for a new open-source benchmark).
        \item The instructions should contain the exact command and environment needed to run to reproduce the results. See the NeurIPS code and data submission guidelines (\url{https://neurips.cc/public/guides/CodeSubmissionPolicy}) for more details.
        \item The authors should provide instructions on data access and preparation, including how to access the raw data, preprocessed data, intermediate data, and generated data, etc.
        \item The authors should provide scripts to reproduce all experimental results for the new proposed method and baselines. If only a subset of experiments are reproducible, they should state which ones are omitted from the script and why.
        \item At submission time, to preserve anonymity, the authors should release anonymized versions (if applicable).
        \item Providing as much information as possible in supplemental material (appended to the paper) is recommended, but including URLs to data and code is permitted.
    \end{itemize}

\item {\bf Experimental setting/details}
    \item[] Question: Does the paper specify all the training and test details (e.g., data splits, hyperparameters, how they were chosen, type of optimizer) necessary to understand the results?
    \item[] Answer: \answerYes{} 
    \item[] Justification: This paper provides comprehensive details on the training and test settings, including data splits, hyperparameters.

    \item[] Guidelines:
    \begin{itemize}
        \item The answer \answerNA{} means that the paper does not include experiments.
        \item The experimental setting should be presented in the core of the paper to a level of detail that is necessary to appreciate the results and make sense of them.
        \item The full details can be provided either with the code, in appendix, or as supplemental material.
    \end{itemize}

\item {\bf Experiment statistical significance}
    \item[] Question: Does the paper report error bars suitably and correctly defined or other appropriate information about the statistical significance of the experiments?
    \item[] Answer: \answerNo{} 
    \item[] Justification: The paper does not provide error bars, confidence intervals, or statistical significance tests for the experiments.
    \item[] Guidelines:
    \begin{itemize}
        \item The answer \answerNA{} means that the paper does not include experiments.
        \item The authors should answer \answerYes{} if the results are accompanied by error bars, confidence intervals, or statistical significance tests, at least for the experiments that support the main claims of the paper.
        \item The factors of variability that the error bars are capturing should be clearly stated (for example, train/test split, initialization, random drawing of some parameter, or overall run with given experimental conditions).
        \item The method for calculating the error bars should be explained (closed form formula, call to a library function, bootstrap, etc.)
        \item The assumptions made should be given (e.g., Normally distributed errors).
        \item It should be clear whether the error bar is the standard deviation or the standard error of the mean.
        \item It is OK to report 1-sigma error bars, but one should state it. The authors should preferably report a 2-sigma error bar than state that they have a 96\% CI, if the hypothesis of Normality of errors is not verified.
        \item For asymmetric distributions, the authors should be careful not to show in tables or figures symmetric error bars that would yield results that are out of range (e.g., negative error rates).
        \item If error bars are reported in tables or plots, the authors should explain in the text how they were calculated and reference the corresponding figures or tables in the text.
    \end{itemize}

\item {\bf Experiments compute resources}
    \item[] Question: For each experiment, does the paper provide sufficient information on the computer resources (type of compute workers, memory, time of execution) needed to reproduce the experiments?
    \item[] Answer: \answerYes{} 
    \item[] Justification: We provide the type of compute workers (GPUs) and the size of memory.
    \item[] Guidelines:
    \begin{itemize}
        \item The answer \answerNA{} means that the paper does not include experiments.
        \item The paper should indicate the type of compute workers CPU or GPU, internal cluster, or cloud provider, including relevant memory and storage.
        \item The paper should provide the amount of compute required for each of the individual experimental runs as well as estimate the total compute. 
        \item The paper should disclose whether the full research project required more compute than the experiments reported in the paper (e.g., preliminary or failed experiments that didn't make it into the paper). 
    \end{itemize}
    
\item {\bf Code of ethics}
    \item[] Question: Does the research conducted in the paper conform, in every respect, with the NeurIPS Code of Ethics \url{https://neurips.cc/public/EthicsGuidelines}?
    \item[] Answer: \answerYes{} 
    \item[] Justification: The research fully adheres to the NeurIPS Code of Ethics.
    \item[] Guidelines:
    \begin{itemize}
        \item The answer \answerNA{} means that the authors have not reviewed the NeurIPS Code of Ethics.
        \item If the authors answer \answerNo, they should explain the special circumstances that require a deviation from the Code of Ethics.
        \item The authors should make sure to preserve anonymity (e.g., if there is a special consideration due to laws or regulations in their jurisdiction).
    \end{itemize}

\item {\bf Broader impacts}
    \item[] Question: Does the paper discuss both potential positive societal impacts and negative societal impacts of the work performed?
    \item[] Answer: \answerYes{} 
    \item[] Justification: The paper discusses both potential positive societal impacts and negative societal impacts of the work performed.
    \item[] Guidelines:
    \begin{itemize}
        \item The answer \answerNA{} means that there is no societal impact of the work performed.
        \item If the authors answer \answerNA{} or \answerNo, they should explain why their work has no societal impact or why the paper does not address societal impact.
        \item Examples of negative societal impacts include potential malicious or unintended uses (e.g., disinformation, generating fake profiles, surveillance), fairness considerations (e.g., deployment of technologies that could make decisions that unfairly impact specific groups), privacy considerations, and security considerations.
        \item The conference expects that many papers will be foundational research and not tied to particular applications, let alone deployments. However, if there is a direct path to any negative applications, the authors should point it out. For example, it is legitimate to point out that an improvement in the quality of generative models could be used to generate Deepfakes for disinformation. On the other hand, it is not needed to point out that a generic algorithm for optimizing neural networks could enable people to train models that generate Deepfakes faster.
        \item The authors should consider possible harms that could arise when the technology is being used as intended and functioning correctly, harms that could arise when the technology is being used as intended but gives incorrect results, and harms following from (intentional or unintentional) misuse of the technology.
        \item If there are negative societal impacts, the authors could also discuss possible mitigation strategies (e.g., gated release of models, providing defenses in addition to attacks, mechanisms for monitoring misuse, mechanisms to monitor how a system learns from feedback over time, improving the efficiency and accessibility of ML).
    \end{itemize}
    
\item {\bf Safeguards}
    \item[] Question: Does the paper describe safeguards that have been put in place for responsible release of data or models that have a high risk for misuse (e.g., pre-trained language models, image generators, or scraped datasets)?
    \item[] Answer: \answerNA{} 
    \item[] Justification: The paper poses no such risks.
    \item[] Guidelines:
    \begin{itemize}
        \item The answer \answerNA{} means that the paper poses no such risks.
        \item Released models that have a high risk for misuse or dual-use should be released with necessary safeguards to allow for controlled use of the model, for example by requiring that users adhere to usage guidelines or restrictions to access the model or implementing safety filters. 
        \item Datasets that have been scraped from the Internet could pose safety risks. The authors should describe how they avoided releasing unsafe images.
        \item We recognize that providing effective safeguards is challenging, and many papers do not require this, but we encourage authors to take this into account and make a best faith effort.
    \end{itemize}

\item {\bf Licenses for existing assets}
    \item[] Question: Are the creators or original owners of assets (e.g., code, data, models), used in the paper, properly credited and are the license and terms of use explicitly mentioned and properly respected?
    \item[] Answer: \answerYes{} 
    \item[] Justification: They are properly credited and the license and terms of use are explicitly mentioned and properly respected.
    \item[] Guidelines:
    \begin{itemize}
        \item The answer \answerNA{} means that the paper does not use existing assets.
        \item The authors should cite the original paper that produced the code package or dataset.
        \item The authors should state which version of the asset is used and, if possible, include a URL.
        \item The name of the license (e.g., CC-BY 4.0) should be included for each asset.
        \item For scraped data from a particular source (e.g., website), the copyright and terms of service of that source should be provided.
        \item If assets are released, the license, copyright information, and terms of use in the package should be provided. For popular datasets, \url{paperswithcode.com/datasets} has curated licenses for some datasets. Their licensing guide can help determine the license of a dataset.
        \item For existing datasets that are re-packaged, both the original license and the license of the derived asset (if it has changed) should be provided.
        \item If this information is not available online, the authors are encouraged to reach out to the asset's creators.
    \end{itemize}

\item {\bf New assets}
    \item[] Question: Are new assets introduced in the paper well documented and is the documentation provided alongside the assets?
    \item[] Answer: \answerYes{} 
    \item[] Justification: The documentation is provided alongside the assets.
    \item[] Guidelines:
    \begin{itemize}
        \item The answer \answerNA{} means that the paper does not release new assets.
        \item Researchers should communicate the details of the dataset\slash code\slash model as part of their submissions via structured templates. This includes details about training, license, limitations, etc. 
        \item The paper should discuss whether and how consent was obtained from people whose asset is used.
        \item At submission time, remember to anonymize your assets (if applicable). You can either create an anonymized URL or include an anonymized zip file.
    \end{itemize}

\item {\bf Crowdsourcing and research with human subjects}
    \item[] Question: For crowdsourcing experiments and research with human subjects, does the paper include the full text of instructions given to participants and screenshots, if applicable, as well as details about compensation (if any)? 
    \item[] Answer: \answerNA{} 
    \item[] Justification: The paper does not involve crowdsourcing nor research with human subjects.
    \item[] Guidelines:
    \begin{itemize}
        \item The answer \answerNA{} means that the paper does not involve crowdsourcing nor research with human subjects.
        \item Including this information in the supplemental material is fine, but if the main contribution of the paper involves human subjects, then as much detail as possible should be included in the main paper. 
        \item According to the NeurIPS Code of Ethics, workers involved in data collection, curation, or other labor should be paid at least the minimum wage in the country of the data collector. 
    \end{itemize}

\item {\bf Institutional review board (IRB) approvals or equivalent for research with human subjects}
    \item[] Question: Does the paper describe potential risks incurred by study participants, whether such risks were disclosed to the subjects, and whether Institutional Review Board (IRB) approvals (or an equivalent approval/review based on the requirements of your country or institution) were obtained?
    \item[] Answer: \answerNA{} 
    \item[] Justification: The paper does not involve crowdsourcing nor research with human subjects.
    \item[] Guidelines:
    \begin{itemize}
        \item The answer \answerNA{} means that the paper does not involve crowdsourcing nor research with human subjects.
        \item Depending on the country in which research is conducted, IRB approval (or equivalent) may be required for any human subjects research. If you obtained IRB approval, you should clearly state this in the paper. 
        \item We recognize that the procedures for this may vary significantly between institutions and locations, and we expect authors to adhere to the NeurIPS Code of Ethics and the guidelines for their institution. 
        \item For initial submissions, do not include any information that would break anonymity (if applicable), such as the institution conducting the review.
    \end{itemize}

\item {\bf Declaration of LLM usage}
    \item[] Question: Does the paper describe the usage of LLMs if it is an important, original, or non-standard component of the core methods in this research? Note that if the LLM is used only for writing, editing, or formatting purposes and does \emph{not} impact the core methodology, scientific rigor, or originality of the research, declaration is not required.
    \item[] Answer: \answerNA{} 
    \item[] Justification: The core method development in this research does not involve LLMs as any important, original, or non-standard components.
    \item[] Guidelines:
    \begin{itemize}
        \item The answer \answerNA{} means that the core method development in this research does not involve LLMs as any important, original, or non-standard components.
        \item Please refer to our LLM policy in the NeurIPS handbook for what should or should not be described.
    \end{itemize}

\end{enumerate}

\end{document}